\documentclass[10pt,twocolumn,letterpaper]{article}

\usepackage{cvpr}              %

\usepackage{graphicx}
\usepackage{amsmath}
\usepackage{amssymb}
\usepackage{booktabs}
\usepackage{amsmath}
\usepackage{booktabs}
\usepackage{etoolbox}
\usepackage{import}
\usepackage{multirow}
\usepackage{siunitx}
\usepackage{tabularx}
\usepackage{xcolor} 
\usepackage[page]{appendix} %

\usepackage[accsupp]{axessibility}  %

\newcommand{\bftab}{\fontseries{b}\selectfont}

\definecolor{MarkusColor}{rgb}{0.44, 0.82, 0.99}

\newcommand{\setcard}[1]{\ensuremath{\vert #1 \vert}}
\newcommand{\images}[0]{\ensuremath{\mathcal{I}}}
\newcommand{\image}[0]{\ensuremath{I}}
\newcommand{\masks}[0]{\ensuremath{\mathcal{M}}}
\newcommand{\mask}[0]{\ensuremath{M}}
\newcommand{\mesh}[0]{\ensuremath{\mathcal{G}}}
\newcommand{\vertices}[0]{\ensuremath{V}}
\newcommand{\edges}[0]{\ensuremath{\mathcal{E}}}
\newcommand{\faces}[0]{\ensuremath{\mathcal{F}}}

\newcommand{\laplacian}[0]{\ensuremath{L}}
\newcommand{\normal}[0]{\ensuremath{\mathbf{n}}}

\newcommand{\colmap}[0]{\textsc{Colmap}}
\newcommand{\idr}[0]{\textsc{IDR}}

\newcommand{\papertitle}[0]{
Multi-View Mesh Reconstruction with Neural Deferred Shading
}

\newcommand{\paperauthors}[0]{
    Markus Worchel$^{1,2}$\thanks{Equal contribution} \hspace{10pt}
    Rodrigo Diaz$^{1,3}$\footnotemark[1] \hspace{10pt}
    Weiwen Hu$^{1}$\\[5pt] %
    Oliver Schreer$^{1}$\hspace{10pt}
    Ingo Feldmann$^{1}$\hspace{10pt}
    Peter Eisert$^{1,4}$ \\[10pt]
    $^1$Fraunhofer HHI \hspace{5pt}
    $^2$TU Berlin\hspace{5pt}
    $^3$Queen Mary University of London\hspace{5pt}
    $^4$HU Berlin
}

\DeclareMathOperator*{\argmin}{arg\,min}
\newtoggle{figpreview}
\togglefalse{figpreview}

\newcommand{\figTeaser}{
\begin{figure}[tb]
    \centering
    \includegraphics[width=\columnwidth]{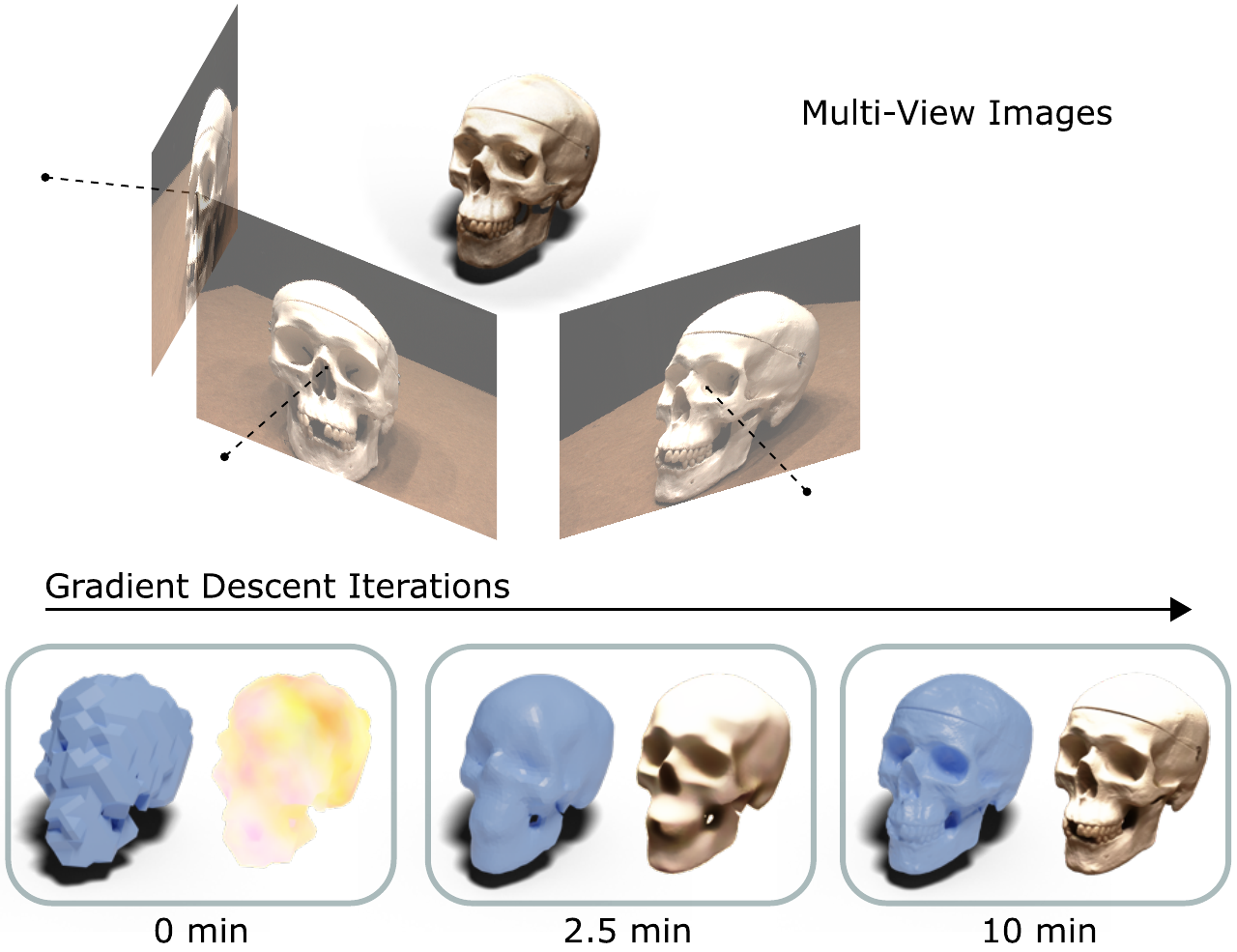}
    \caption{We reconstruct an object from images by simultaneously deforming a triangle mesh and optimizing a neural shader, comparing the renderings to the input images.}
    \label{fig:teaser}
\end{figure}
}

\newcommand{\figOverview}{
\begin{figure*}[ht]
    \centering
    \includegraphics[width=\textwidth]{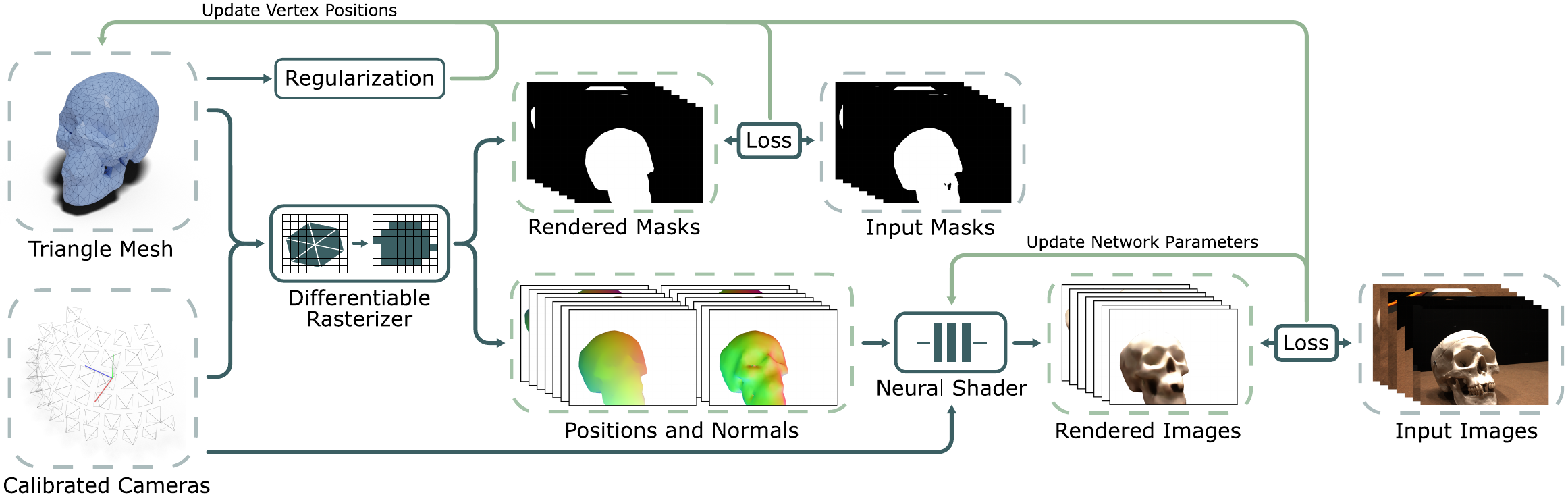}
    \caption{Overview of our optimization procedure. We rasterize the triangle mesh and shade the results with a neural network to synthesize an image for each input camera view. The shader is updated based on the difference between rendered and input images, whereas the mesh vertices are also updated based on the silhouette and a geometric regularization term. We optimize using gradient descent.}
    \label{fig:pipeline}
\end{figure*}
}

\newcommand{\figShaderArchitecture}{
\begin{figure}[h]
    \centering
    \includegraphics[width=0.8\columnwidth]{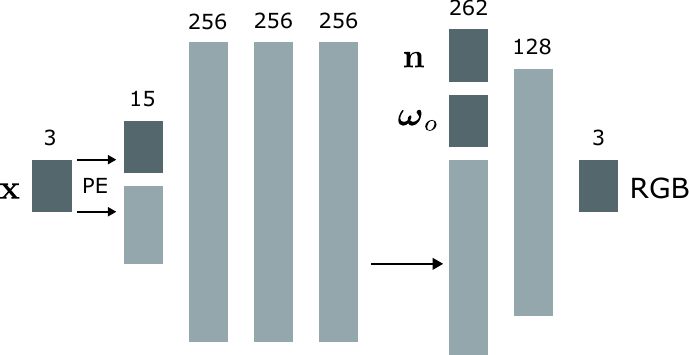}
    \caption{Architecture of the neural shader. The position $\mathbf{x}$ is transformed by a positional encoding (PE)~\cite{Tancik:2020:FourierFeat} and processed by 3 fully-connected layers. The resulting feature vector is concatenated with the surface normal $\mathbf{n}$ and the view direction $\boldsymbol{\omega}_o$ and processed by the last two layers, yielding a color value. We use ReLU activations for the hidden layers and a sigmoid activation for the last layer.}
    \label{fig:shader_architecture}
\end{figure}
}

\iftoggle{figpreview}{%
    \newcommand{\reconstructiondtuimage}[1]{\raisebox{-0.5\height}{\includegraphics[width=0.23\columnwidth, trim=37.5 30 45 35]{images/reconstruction_dtu/preview/#1}}}
}{%
    \newcommand{\reconstructiondtuimage}[1]{\raisebox{-0.5\height}{\includegraphics[width=0.23\columnwidth, trim=75 60 90 70]{images/reconstruction_dtu/medium/#1}}}
}

\newcommand{\figReconstructionDTU}{
\begin{figure*}[htb]
    \centering
    \setlength{\tabcolsep}{1.8pt}
    \begin{tabular}{ccccccccc}
        \reconstructiondtuimage{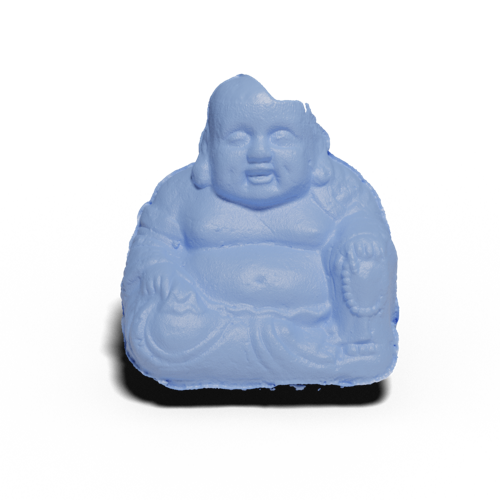} &
        \reconstructiondtuimage{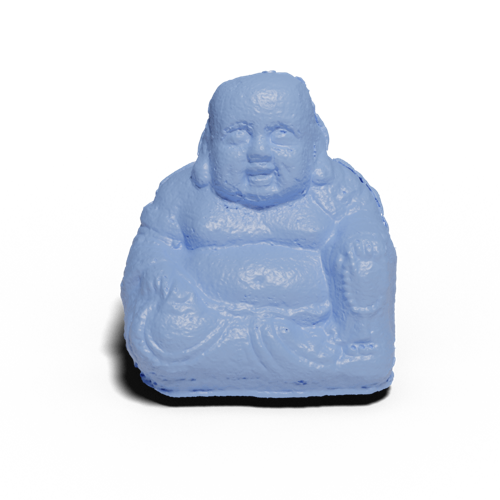} &
        \reconstructiondtuimage{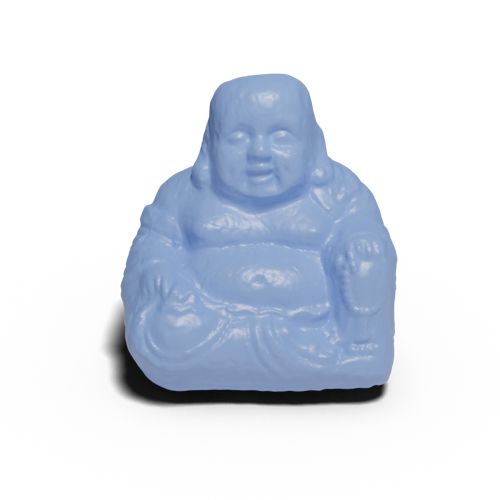} & 
        \reconstructiondtuimage{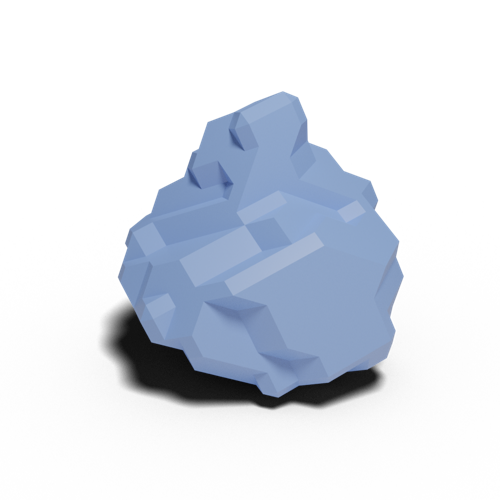} &
        \reconstructiondtuimage{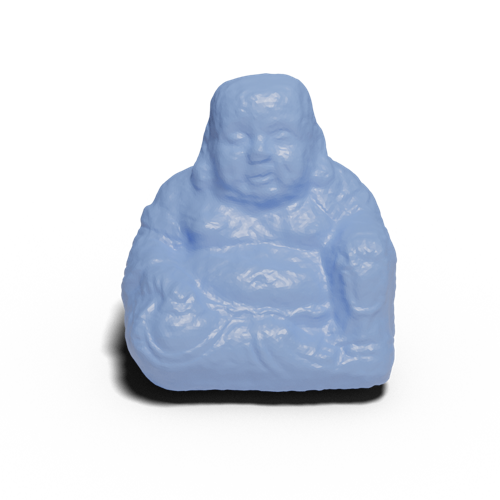} &
        \quad \reconstructiondtuimage{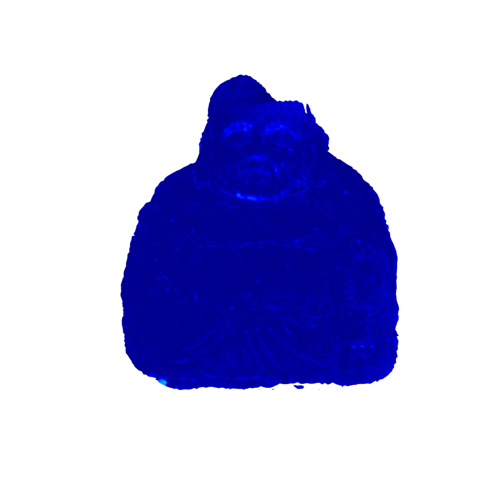} & 
        \reconstructiondtuimage{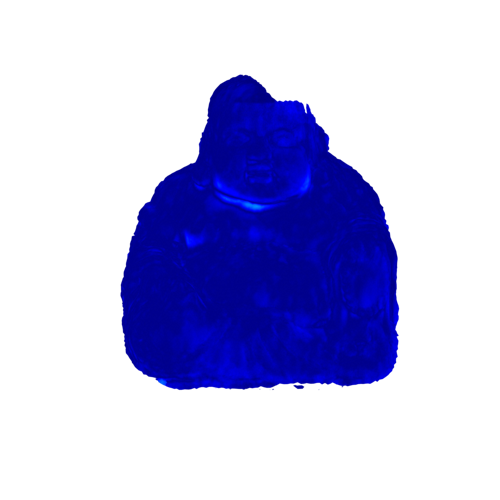} & 
        \reconstructiondtuimage{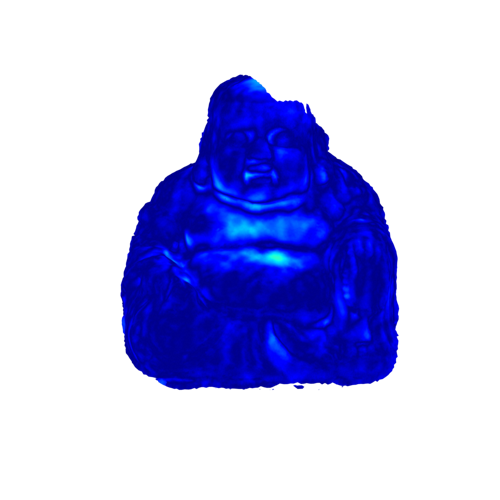} & 
        \multirow{2}{*}{\includegraphics[height=0.4\columnwidth]{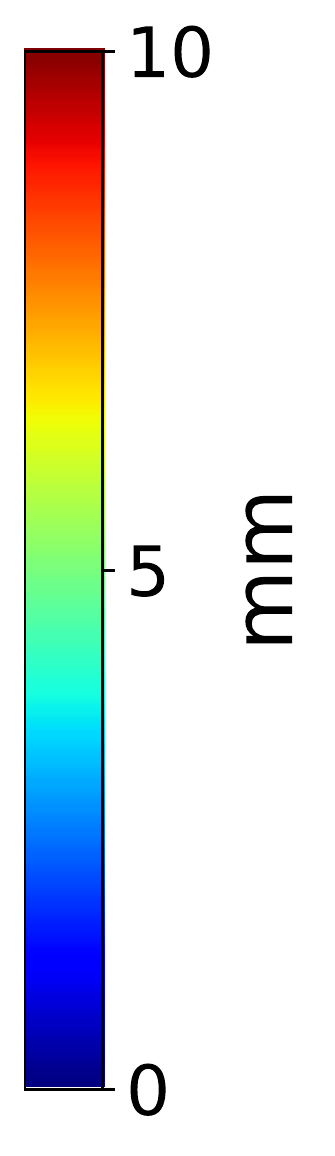}} \\
        \reconstructiondtuimage{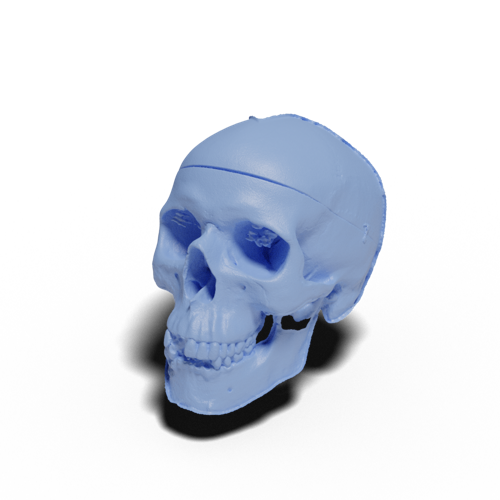} &
        \reconstructiondtuimage{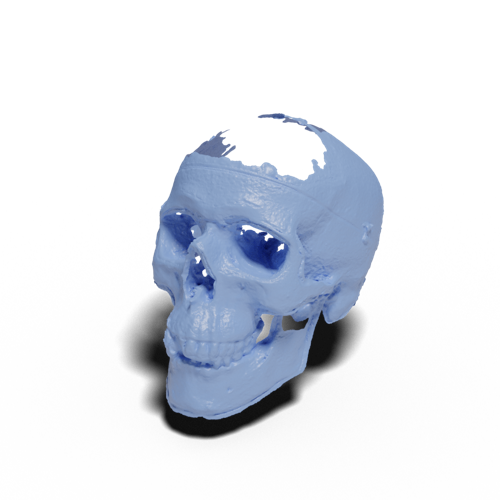} &
        \reconstructiondtuimage{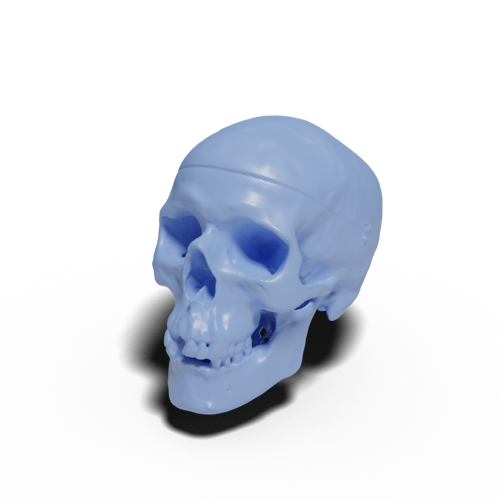} & 
        \reconstructiondtuimage{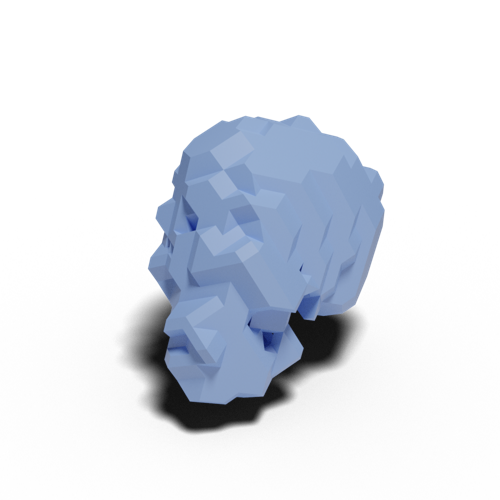} &
        \reconstructiondtuimage{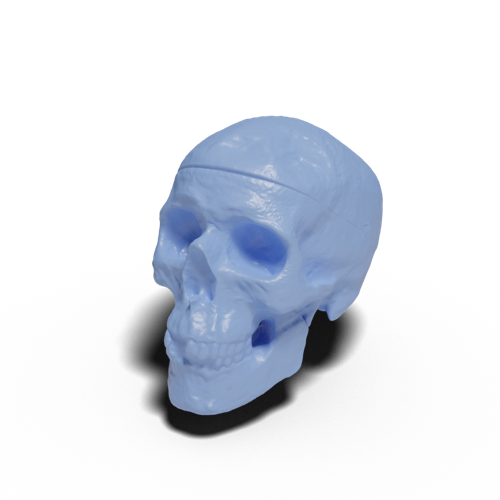} &
        \quad \reconstructiondtuimage{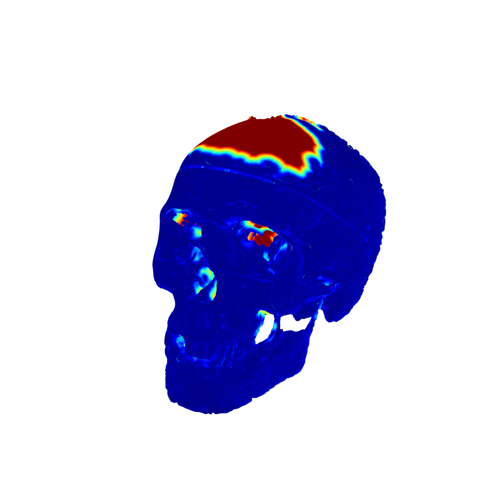} & 
        \reconstructiondtuimage{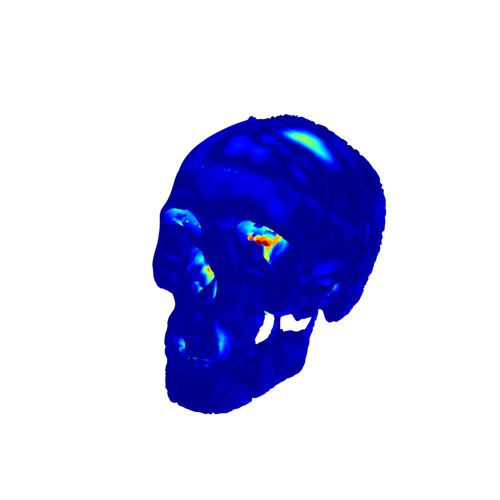} & 
        \reconstructiondtuimage{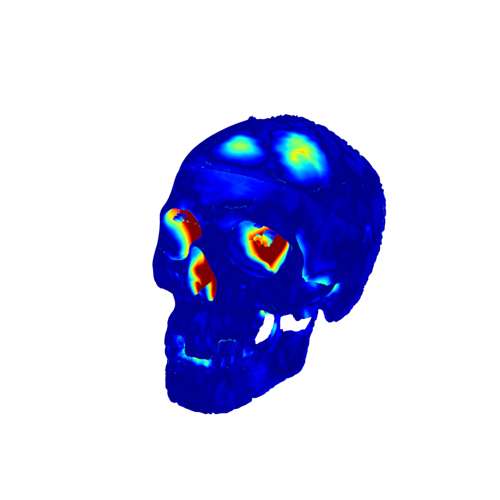} & \\
        \footnotesize{Reference} & \footnotesize{\colmap} & \footnotesize{IDR} & \footnotesize{Ours (Initial)} & \footnotesize{Ours} & 
        \quad \footnotesize{\colmap} & \footnotesize{IDR} & \footnotesize{Ours} &
    \end{tabular}
    \caption{Qualitative comparison on the DTU dataset. Left: Reference geometry and reconstruction results. Right: Point-to-mesh distance between the reference scan and the reconstructed mesh.}
    \label{fig:reconstruction_dtu}
\end{figure*}
}

\newcommand{\tabReconstructionDTU}{
\begin{table}[t]
	\centering
	\caption{Quantitative results for multi-view reconstruction of objects from the DTU dataset. Chamfer scores are in \emph{millimeters} and the \colmap{} runtime is for the trim7 configuration.}
	\label{table:reconstruction_dtu}
	\resizebox{\columnwidth}{!}{%
	\begin{tabular}{@{}lrrrrrr@{}} 
		\toprule
		\multirow{3}{*}{Scan} & \multicolumn{2}{c}{\colmap{}~\cite{Schoenberger:2016:sfm, Schoenberger:2016:mvs}} & \multicolumn{2}{c}{IDR~\cite{Yariv:2020:multiview}} & \multicolumn{2}{c}{NDS~(Ours)} \\
		& \multicolumn{2}{c}{\footnotesize trim7 (trim0)} & & \\
		\cmidrule(lr){2-3} \cmidrule(lr){4-5} \cmidrule(l){6-7}
		& $\downarrow$ Chamfer-L1 & $\downarrow$ time [min] & $\downarrow$ Chamfer-L1 & $\downarrow$ time [min] & $\downarrow$ Chamfer-L1 & $\downarrow$ time [min] \\
		\midrule
        24 & {\bftab 0.45} (0.81) & 66.81 & 1.58 & 551.83 & 4.24 & \bftab 12.24 \\
        37 & {\bftab 0.90} (2.03) & 81.19 & 2.06 & 566.13 & 5.25 & \bftab 7.56 \\
        40 & {\bftab 0.36} (0.75) & 65.47 & 0.75 & 550.19 & 1.30 & \bftab 10.69 \\
        55 & {\bftab 0.36} (1.20) & 64.71 & 0.43 & 565.27 & 0.53 & \bftab 7.54 \\
        63 & {\bftab 0.90} (1.75) & 75.62 & 1.06 & 553.07 & 2.47 & \bftab 6.18 \\
        65 & 0.94 (1.55) & 62.62 & \bftab 0.79 & 568.41 & 1.22 & \bftab 9.95 \\
        69 & {\bftab 0.53} (1.02) & 86.77 & 0.68 & 557.58 & 1.35 & \bftab 9.64 \\
        83 & {\bftab 1.16} (3.03) & 77.96 & 1.38 & 745.18 & 1.59 & \bftab 4.76 \\
        97 & {\bftab 1.08} (1.42) & 50.89 & 1.17 & 743.11 & 2.77 & \bftab 7.32 \\
        105 & {\bftab 0.63} (1.96) & 51.59 & 0.88 & 742.31 & 1.15 & \bftab 7.00 \\
        106 & {\bftab 0.48} (0.99) & 108.01 & 0.63 & 735.23 & 1.02 & \bftab 7.41 \\
        110 & {\bftab 0.58} (1.33) & 85.87 & 0.99 & 752.94 & 3.18 & \bftab 6.68 \\
        114 & {\bftab 0.31} (0.50) & 88.39 & 0.37 & 730.50 & 0.62 & \bftab 7.99 \\
        118 & {\bftab 0.44} (0.78) & 105.75 & 0.50 & 748.60 & 1.65 & \bftab 6.77 \\
        122 & {\bftab 0.43} (1.17) & 80.80 & 0.52 & 747.90 & 0.91 & \bftab 5.76 \\
        \midrule
        mean & {\bftab 0.64} (1.35) & 76.83 & 0.92 & 657.22 & 1.95 & \bftab 7.83 \\
		\bottomrule
	\end{tabular}
	}
\end{table}
}

\iftoggle{figpreview}{%
    \newcommand{\reconstructiondtuequaltimeimage}[1]{\raisebox{-0.5\height}{\includegraphics[width=0.29\columnwidth, trim=10 35 15 40]{images/reconstruction_dtu_equal_time/preview/#1}}}
}{%
    \newcommand{\reconstructiondtuequaltimeimage}[1]{\raisebox{-0.5\height}{\includegraphics[width=0.29\columnwidth, trim=20 20 30 80]{images/reconstruction_dtu_equal_time/medium/#1}}}
}

\newcommand{\figReconstructionDTUEqualTime}{
\begin{figure}[htb]
    \centering
    \setlength{\fboxsep}{0pt}
    \setlength{\tabcolsep}{2pt}
    \begin{tabular}{@{}cccc@{}}
        \rotatebox[origin=c]{90}{\scriptsize{\textsc{IDR}}} &
        \reconstructiondtuequaltimeimage{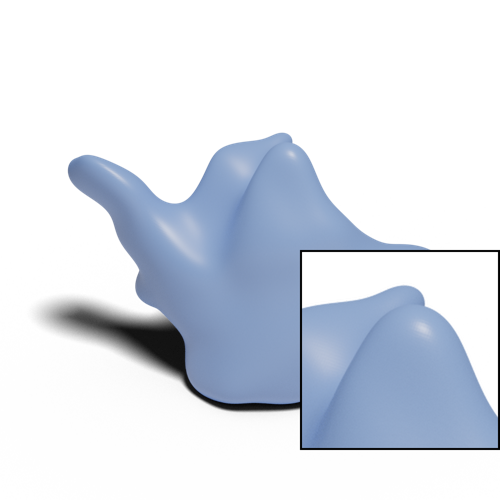} & 
        \reconstructiondtuequaltimeimage{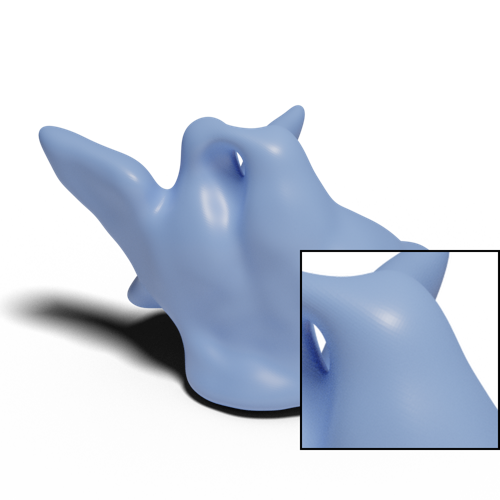} &
        \reconstructiondtuequaltimeimage{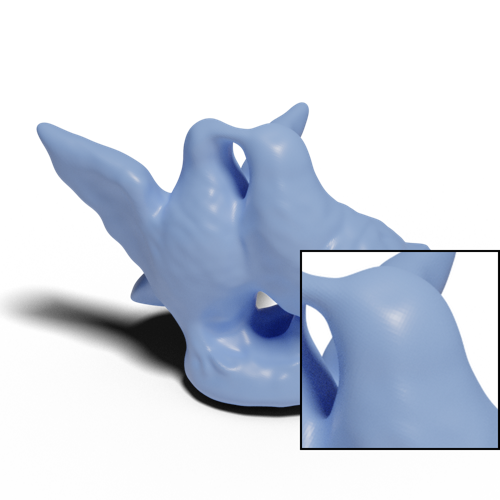} \\
        \rotatebox[origin=c]{90}{\scriptsize{Ours}} &
        \reconstructiondtuequaltimeimage{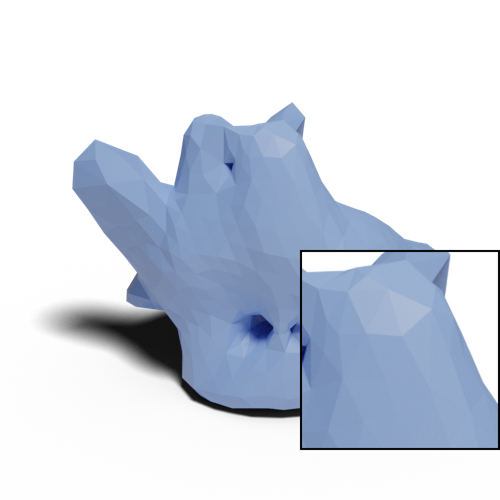} &
        \reconstructiondtuequaltimeimage{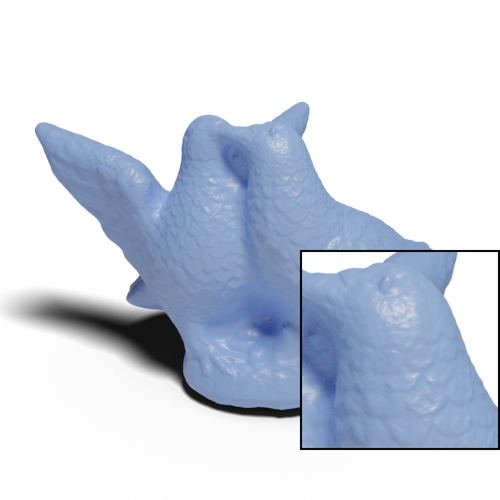} & \\
        &
        \footnotesize{1 min} & 
        \footnotesize{10 min} &
        \footnotesize{50 min} \\
    \end{tabular}
    \caption{Equal time reconstruction. We consider our result converged after 10 minutes. Even after 50 minutes, \idr{} lacks details that are present in our reconstruction (\eg{}, feathers and eyes).}
    \label{fig:reconstruction_dtu_equal_time}
\end{figure}
}

\newcommand{\tabRuntime}{
\begin{table}[t]
	\centering
	\caption{Average runtime of one gradient descent iteration of \idr{} and our method, decomposed hierarchically. Geometry rendering time excludes shading and corresponds to ray marching 2048 pixels for \idr{} and rasterizing to 1.9 million pixels in our case.}
	\label{table:runtime}
	\resizebox{\columnwidth}{!}{%
    \begin{tabular}{@{}lrlrl@{}}
        \toprule
        & \multicolumn{2}{c}{IDR} & \multicolumn{2}{c}{NDS (Ours)} \\
        \cmidrule(lr){2-3} \cmidrule(l){4-5}
        & time [s] & share & time [s] & share  \\
        \midrule
        Gradient descent iteration & 0.3577 & 100\,\% & 0.1561 & 100\,\% \\
        \hspace*{0.1em} \rotatebox[origin=c]{180}{$\Lsh$} Geometry rendering & 0.2099 & \hspace*{0.01em} \rotatebox[origin=c]{180}{$\Lsh$} 58.7\,\% & 0.0034 & \hspace*{0.01em} \rotatebox[origin=c]{180}{$\Lsh$} 2.2\,\% \\
        \hspace*{1.2em} \rotatebox[origin=c]{180}{$\Lsh$} SDF evaluation & 0.1472 & \hspace*{1em} \rotatebox[origin=c]{180}{$\Lsh$} 70.1\,\% & -- & \\
        \bottomrule
    \end{tabular}
	}
\end{table}
}

\newcommand{\meshrefinementmosaicimage}[1]{\fbox{\includegraphics[width=0.12\columnwidth, trim=0 0 0 35, clip]{images/mesh_refinement_hbr/#1}}}

\newcommand{\imgMeshRefinementMosaic}[1]{
    \setlength{\fboxsep}{0pt}
    \setlength{\tabcolsep}{2pt}
    \begin{tabular}{@{}cc@{}}
    \meshrefinementmosaicimage{#1_image_0.png} & \meshrefinementmosaicimage{#1_image_1.png} \\
    \meshrefinementmosaicimage{#1_image_2.png} & \meshrefinementmosaicimage{#1_image_3.png} \\
    \end{tabular}
}

\iftoggle{figpreview}{%
    \newcommand{\meshrefinementimage}[1]{\raisebox{-0.45\height}{\includegraphics[width=0.28\columnwidth, trim=30 20 30 0]{images/mesh_refinement_hbr/#1}}}
}{%
    \newcommand{\meshrefinementimage}[1]{\raisebox{-0.45\height}{\includegraphics[width=0.28\columnwidth, trim=60 40 60 0]{images/mesh_refinement_hbr/medium/#1}}}
}

\newcommand{\figMeshRefinement}{
\begin{figure}[htb]
    \centering
    \setlength{\fboxsep}{0pt}
    \setlength{\tabcolsep}{7pt}
    \begin{tabular}{@{}ccc@{}}
        \imgMeshRefinementMosaic{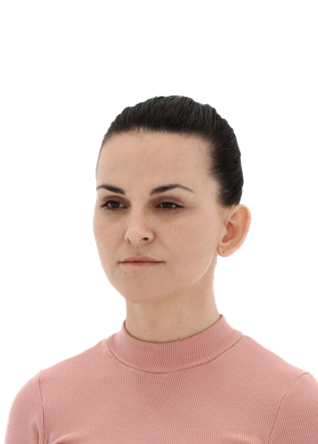} &
        \meshrefinementimage{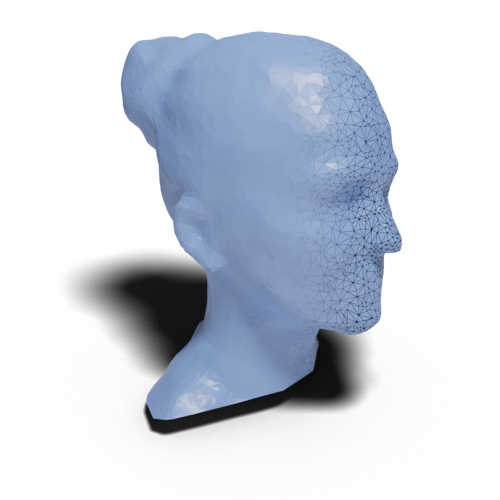} & 
        \meshrefinementimage{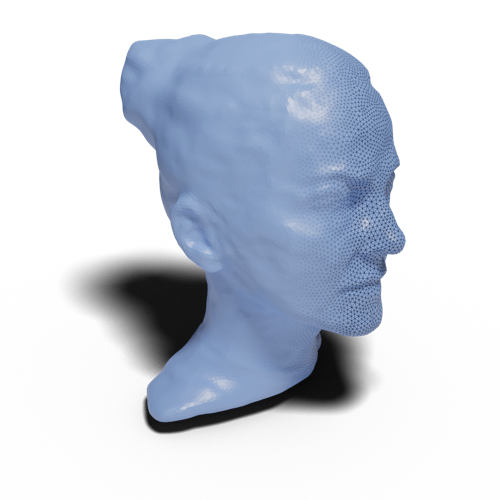} \\
        \imgMeshRefinementMosaic{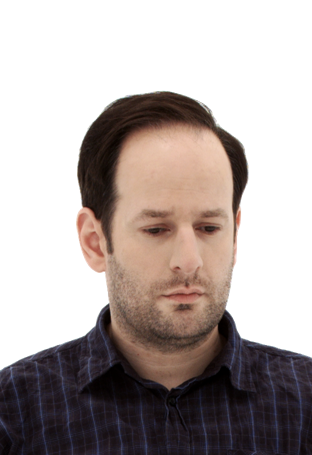} & 
        \meshrefinementimage{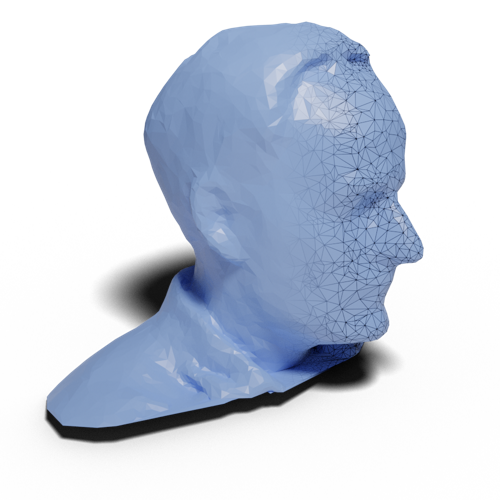} & 
        \meshrefinementimage{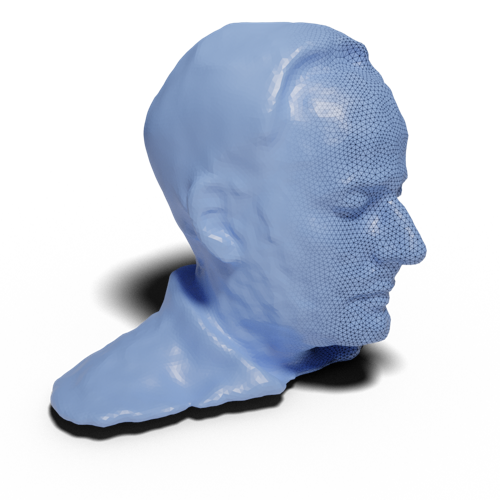} \\
        \footnotesize{Images} &
        \footnotesize{Initial} & 
        \footnotesize{Refined} \\
    \end{tabular}
    
    \caption{Refining meshes from an established multi-view reconstruction pipeline. Four of the 32 images are shown.}
    \label{fig:mesh_refinement}
\end{figure}
}

\iftoggle{figpreview}{%
    \newcommand{\shaderviewsynthimage}[1]{\raisebox{-0.5\height}{\includegraphics[width=0.22\columnwidth, trim=105 0 50 0, clip]{images/view_synthesis/#1}}}
}{%
    \newcommand{\shaderviewsynthimage}[1]{\raisebox{-0.5\height}{\includegraphics[width=0.22\columnwidth, trim=105 0 50 0, clip]{images/view_synthesis/#1}}}
}
    
\newcommand{\shaderviewsynthimagee}[1]{\raisebox{-0.5\height}{\includegraphics[width=0.22\columnwidth, trim=105 50 100 30, clip]{images/view_synthesis/#1}}}
    
\newcommand{\figShaderViewSynth}{
\begin{figure}[htb]
    \centering
    \setlength{\tabcolsep}{5pt}
    \setlength{\fboxsep}{0pt}
    \begin{tabular}{@{}cccc@{}}
        \shaderviewsynthimage{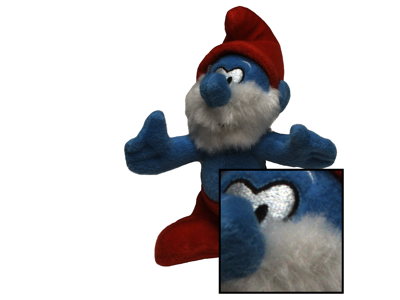} &
        \shaderviewsynthimage{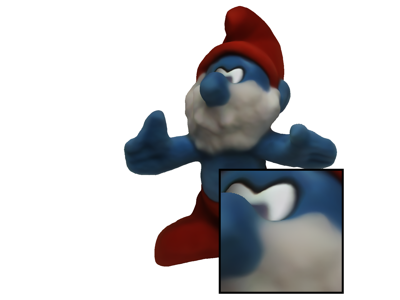} &
        \shaderviewsynthimage{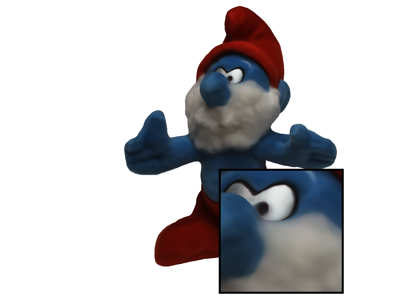} &
        \shaderviewsynthimagee{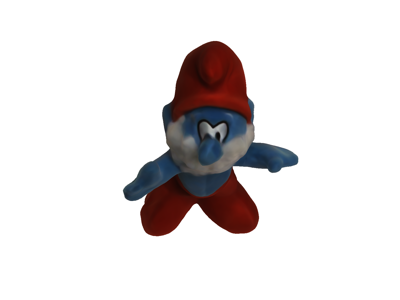}\\
        \footnotesize{Original} &
        \footnotesize{Rendered} & 
        \footnotesize{Rendered} &
        \footnotesize{Rendered\vspace*{-0.4\baselineskip}}\\
        &
        &
        \scriptsize{(fixed mesh)} &
        \scriptsize{(new view)} \\
    \end{tabular}
    \caption{View synthesis for an input view and a new view. By fixing the mesh after reconstruction and continuing shader optimization, we can further improve the view synthesis results.}
    \label{fig:shader_view_synth}
\end{figure}
}

\iftoggle{figpreview}{%
    \newcommand{\shaderpcaimage}[1]{\raisebox{-0.5\height}{\includegraphics[width=0.3\columnwidth, trim=105 0 50 0, clip]{images/shader_analysis/#1}}}
}{%
    \newcommand{\shaderpcaimage}[1]{\raisebox{-0.5\height}{\includegraphics[width=0.3\columnwidth, trim=105 0 50 0, clip]{images/shader_analysis/#1}}}
}
    
\newcommand{\figShaderPca}{
\begin{figure}[htb]
    \centering
    \setlength{\tabcolsep}{5pt}
    \begin{tabular}{@{}ccc@{}}
        \shaderpcaimage{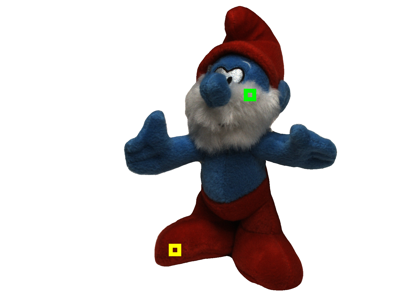} &
        \shaderpcaimage{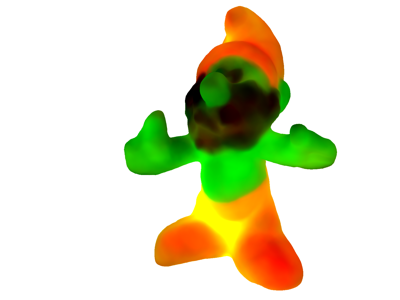} &
        \shaderpcaimage{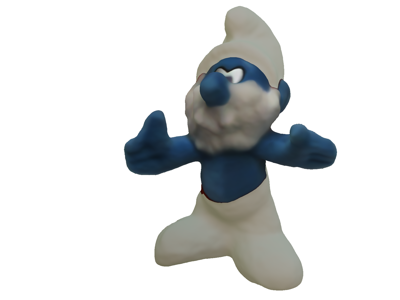} \\
        \footnotesize{Image} &
        \footnotesize{Feature Projection} &
        \footnotesize{Material Edit} \\
    \end{tabular}
    \caption{A principle component analysis of the shader's positional features reveals a connection to materials. We show the projection to the two largest principle components. Replacing features in the positional latent space before computing the view-dependent color enables simple material editing. The feature vector at the yellow square (pants material) determines regions that are replaced with the feature vector at the green square (beard material).}
    \label{fig:shader_pca}
\end{figure}
}

\newcommand{\ablationinitimage}[1]{\raisebox{-0.5\height}{\includegraphics[width=0.21\columnwidth, trim=60 50 60 100]{images/ablation_initial_mesh/#1}}}

\newcommand{\figAblationInitialGeometry}{
\begin{figure}[htb]
    \centering
    \setlength{\fboxsep}{0pt}
    \setlength{\tabcolsep}{2pt}
    \begin{tabular}{@{}ccccc@{}}
        &
        \scriptsize{6.43 min} & 
        \scriptsize{7.82 min} &
        \scriptsize{8.90 min} & 
        \scriptsize{6.96 min} \\
        \rotatebox[origin=c]{90}{\scriptsize{Initial}} &
        \ablationinitimage{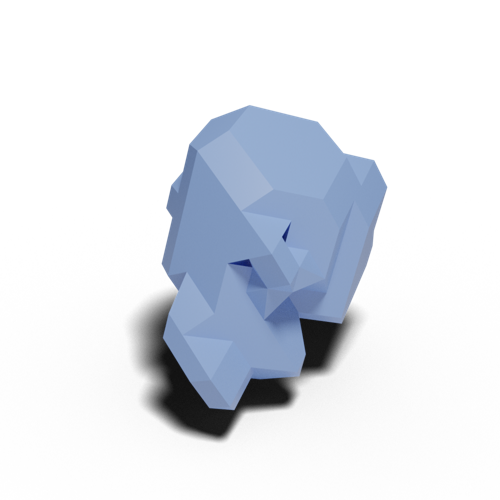} & 
        \ablationinitimage{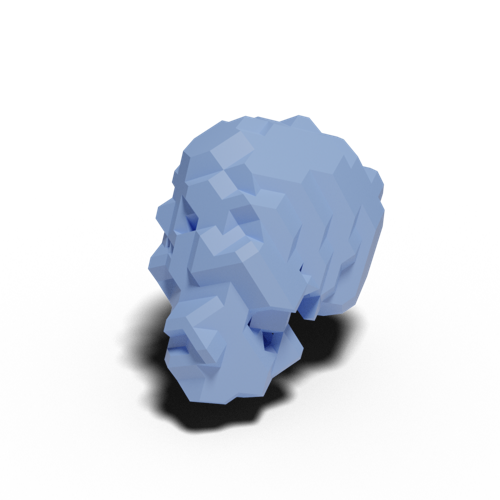} &
        \ablationinitimage{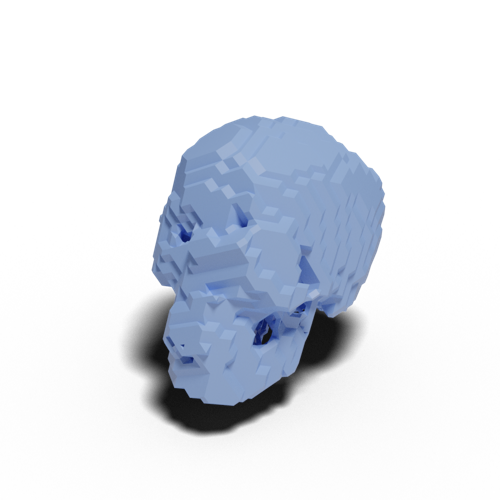} &
        \ablationinitimage{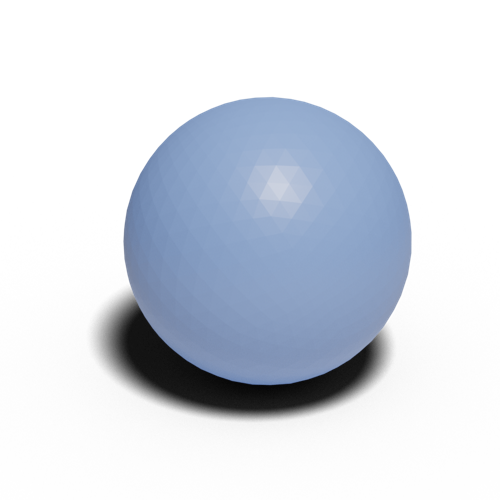} \\
        &
        \scriptsize{540 faces} & 
        \scriptsize{2552 faces} &
        \scriptsize{11366 faces} & 
        \scriptsize{2048 faces} \\
        \rotatebox[origin=c]{90}{\scriptsize{Final}} &
        \ablationinitimage{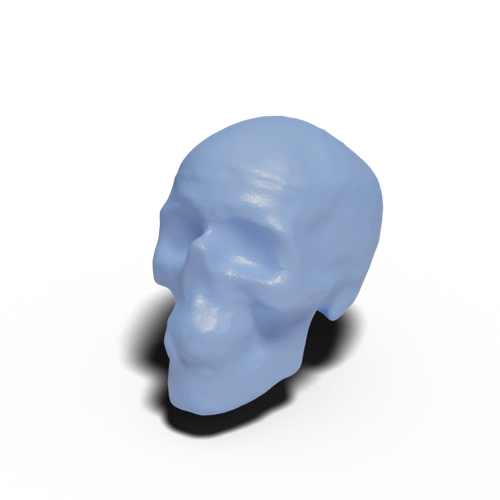} & 
        \ablationinitimage{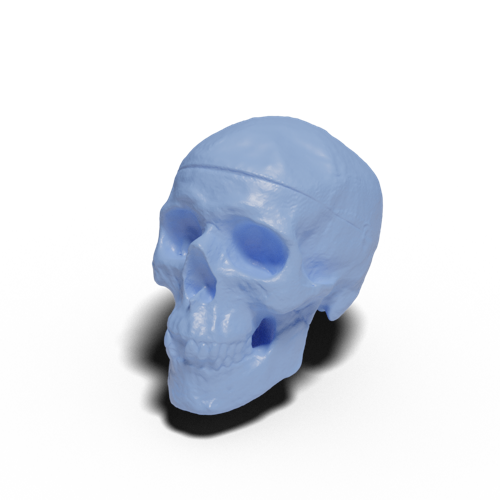} &
        \ablationinitimage{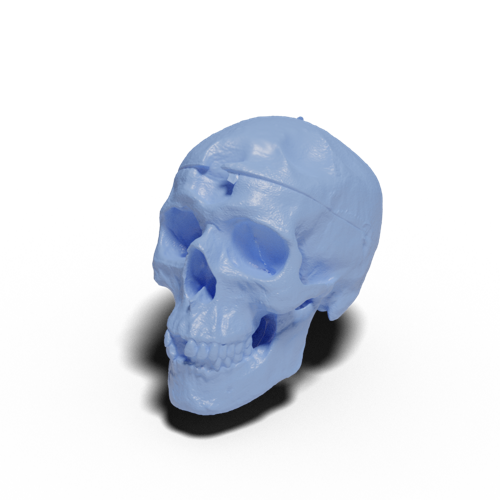} &
        \ablationinitimage{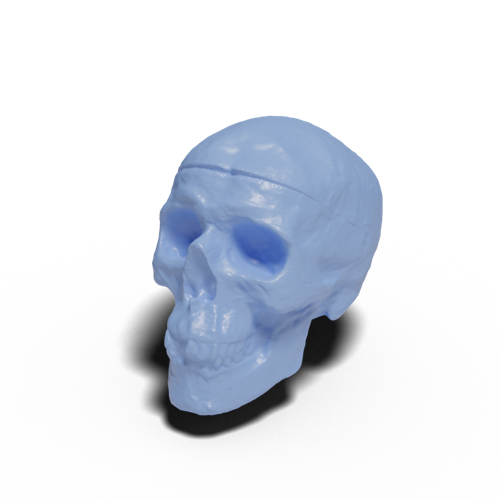} \\
        &
        \scriptsize{31106 faces} & 
        \scriptsize{141278 faces} &
        \scriptsize{604146 faces} & 
        \scriptsize{124242 faces} \\
        &
        \footnotesize{VH $16^3$} & 
        \footnotesize{VH $32^3$ (Ours)} &
        \footnotesize{VH $64^3$} & 
        \footnotesize{Sphere}
    \end{tabular}
    \caption{Reconstruction results for different initial meshes. We vary the resolution of the grid used to build the visual hull (VH) and also start from a sphere.}
    \label{fig:reconstruction_ablation_initial_geometry}
\end{figure}
}

\newcommand{\tabRuntimeExtended}{
\begin{table}[t]
	\centering
	\caption{Full hierarchical runtime decomposition of our method for DTU scan 122 (``Owl") with 2000 iterations. Invocations of the neural shader are included in the shading term. Some one-time operations before to the optimization loop are not explicitly listed (\eg{} initialization of the neural network, optimizers, and the rasterizer). In this experiment, we pre-computed the initial mesh. Computing the visual hull takes roughly 0.5 seconds.}
	\label{table:runtime_extended_sup}
    \begin{tabular}{@{}lrl@{}}
        \toprule
        Operation & Time [s]  \\
        \midrule
        Total & 341.03 \hspace*{3em} \\
        \hspace*{0.1em} \rotatebox[origin=c]{180}{$\Lsh$} Reading data & 20.54 \hspace*{2em} \\
        \hspace*{1.2em} \rotatebox[origin=c]{180}{$\Lsh$} Images & 20.50 \hspace*{1em} \\
        \hspace*{1.2em} \phantom{\rotatebox[origin=c]{180}{$\Lsh$}} Mesh & 0.04 \hspace*{1em} \\
        \hspace*{0.1em} \phantom{\rotatebox[origin=c]{180}{$\Lsh$}} Optimization loop & 312.16 \hspace*{2em} \\
        \hspace*{1.2em} \rotatebox[origin=c]{180}{$\Lsh$} Applying vertex offsets & 2.26 \hspace*{1em} \\
        \hspace*{1.2em} \phantom{\rotatebox[origin=c]{180}{$\Lsh$}} Sampling views & 0.63 \hspace*{1em} \\
        \hspace*{1.2em} \phantom{\rotatebox[origin=c]{180}{$\Lsh$}} Creating g-buffers & 11.49 \hspace*{1em} \\
        \hspace*{2.3em} \rotatebox[origin=c]{180}{$\Lsh$} Rasterization & 6.86 \\
        \hspace*{2.3em} \phantom{\rotatebox[origin=c]{180}{$\Lsh$}} Interpolating coverage & 1.50 \\
        \hspace*{2.3em} \phantom{\rotatebox[origin=c]{180}{$\Lsh$}} Interpolating positions & 1.45 \\
        \hspace*{2.3em} \phantom{\rotatebox[origin=c]{180}{$\Lsh$}} Interpolating normals & 1.43 \\
        \hspace*{1.2em} \phantom{\rotatebox[origin=c]{180}{$\Lsh$}} Computing objective function & 135.63 \hspace*{1em} \\
        \hspace*{2.3em} \rotatebox[origin=c]{180}{$\Lsh$} Silhouette & 0.73 \\
        \hspace*{2.3em} \phantom{\rotatebox[origin=c]{180}{$\Lsh$}} Shading & 128.50 \\
        \hspace*{2.3em} \phantom{\rotatebox[origin=c]{180}{$\Lsh$}} Laplacian & 3.58 \\
        \hspace*{2.3em} \phantom{\rotatebox[origin=c]{180}{$\Lsh$}} Normal consistency & 1.37 \\
        \hspace*{2.3em} \phantom{\rotatebox[origin=c]{180}{$\Lsh$}} Aggregating terms & 1.24 \\
        \hspace*{1.2em} \phantom{\rotatebox[origin=c]{180}{$\Lsh$}} Computing gradients & 129.12 \hspace*{1em} \\
        \hspace*{1.2em} \phantom{\rotatebox[origin=c]{180}{$\Lsh$}} Performing descent step & 5.79 \hspace*{1em} \\
        \hspace*{1.2em} \phantom{\rotatebox[origin=c]{180}{$\Lsh$}} Remeshing & 23.28 \hspace*{1em} \\
        \bottomrule
    \end{tabular}
\end{table}
}

\newcommand{\failurecasesimage}[1]{\raisebox{-0.5\height}{\includegraphics[width=0.3\columnwidth, trim=0 0 200 0, clip]{images/failure_cases/#1}}}
    
\newcommand{\figFailureCases}{
\begin{figure*}[htb]
    \centering
    \renewcommand{\arraystretch}{1.5}
    \setlength{\tabcolsep}{2pt}
    \begin{tabular}{@{}cccccc@{}}
        \failurecasesimage{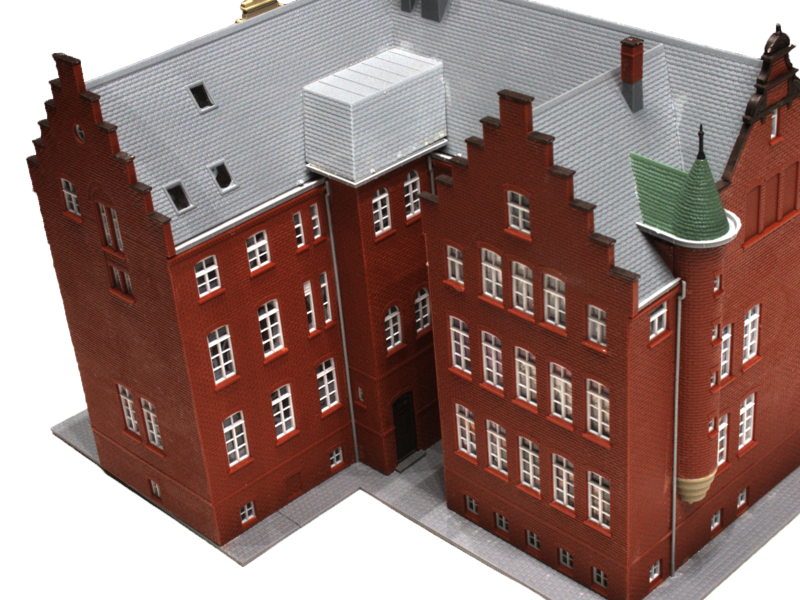} &
        \failurecasesimage{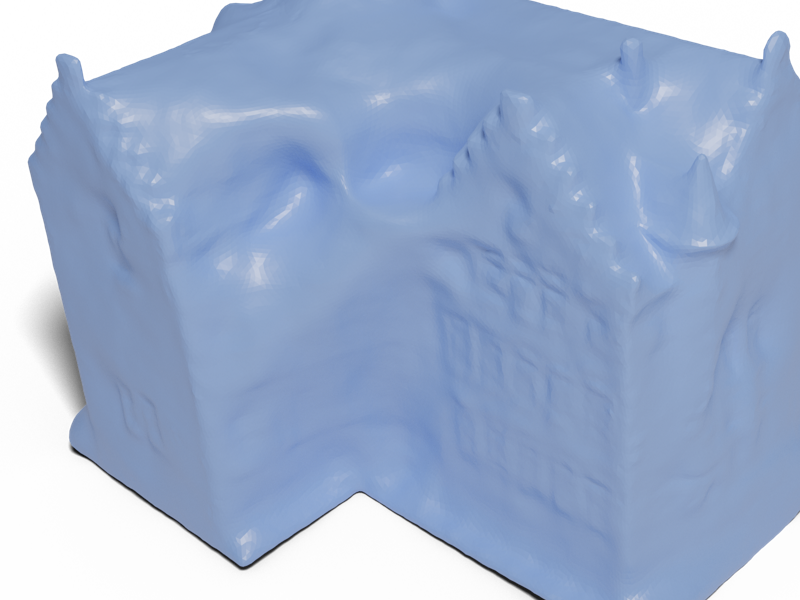} &
        \quad \failurecasesimage{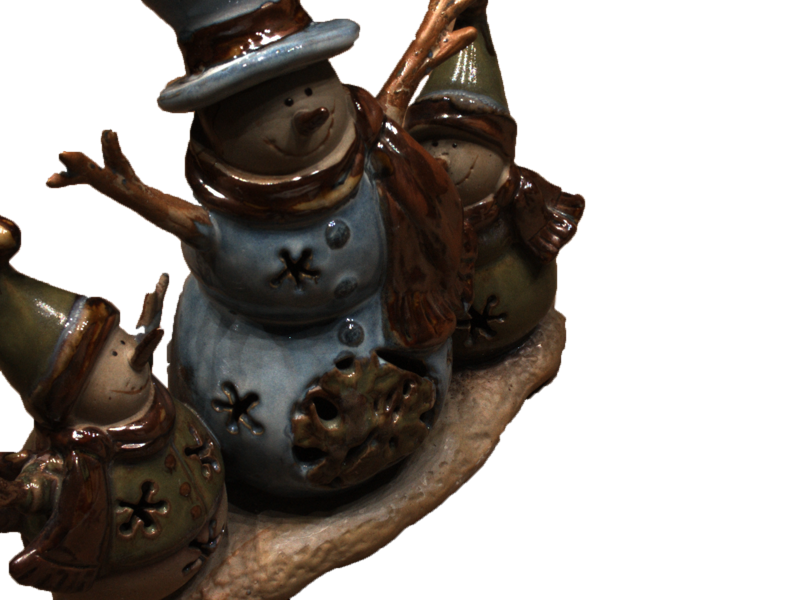} &
        \failurecasesimage{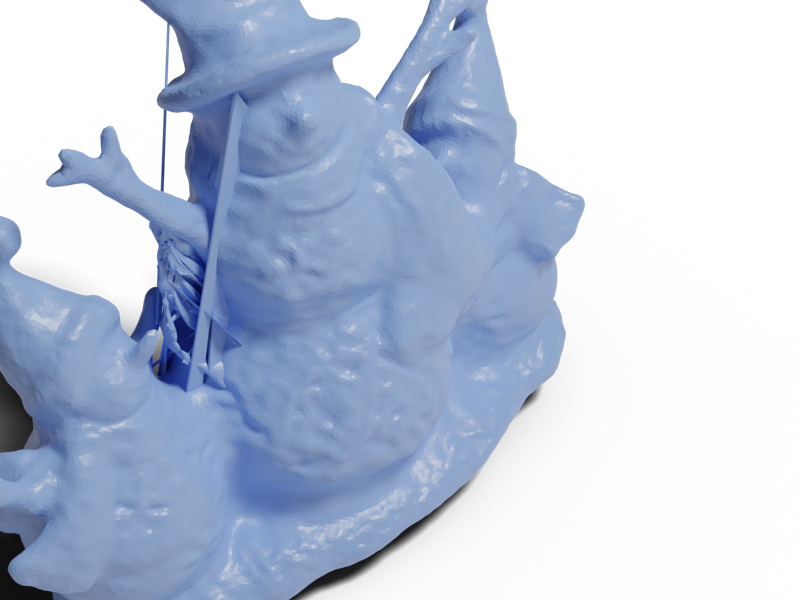} &
        \quad \failurecasesimage{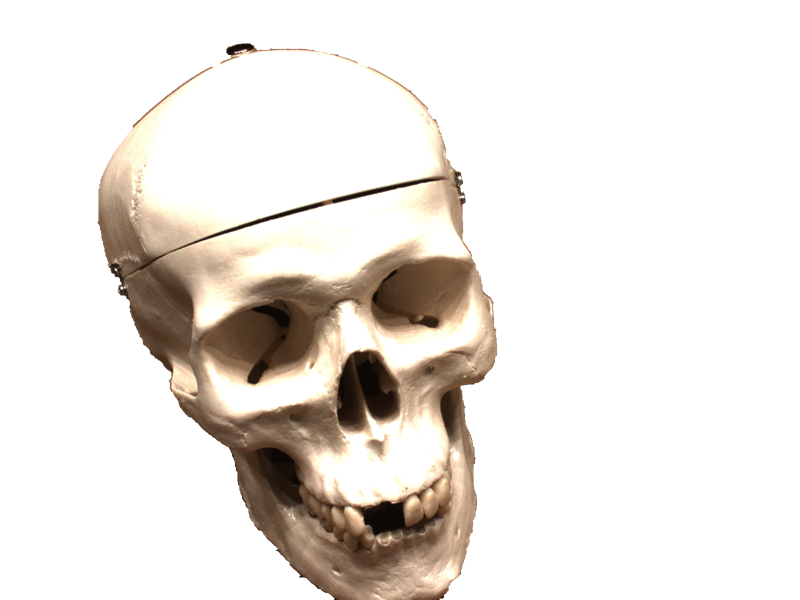} &
        \failurecasesimage{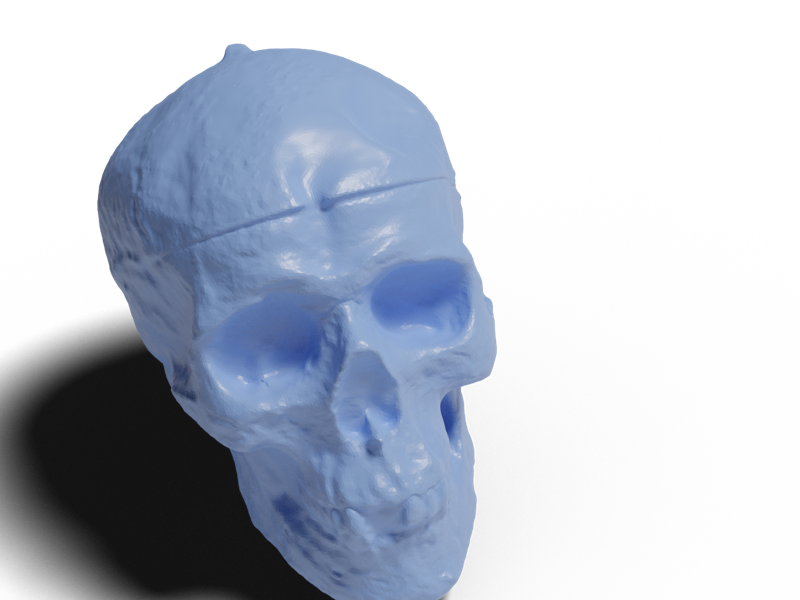} \\
        \multicolumn{2}{c}{\footnotesize{Deep concavities}} & 
        \multicolumn{2}{c}{\quad \footnotesize{Failed remeshing}} &
        \multicolumn{2}{c}{\quad\footnotesize{Overexposed regions}} \\
    \end{tabular}
    \caption{Failure cases of our reconstruction. }
    \label{fig:failure_cases_sup}
\end{figure*}
}

\newcommand{\visualhullimage}[1]{\raisebox{-0.5\height}{\includegraphics[width=0.3\columnwidth, trim=60 30 30 30]{images/visual_hull/#1}}}
    
\newcommand{\figVisualHull}{
\begin{figure}[htb]
    \centering
    \renewcommand{\arraystretch}{1.5}
    \setlength{\tabcolsep}{2pt}
    \begin{tabular}{@{}ccc@{}}
        \visualhullimage{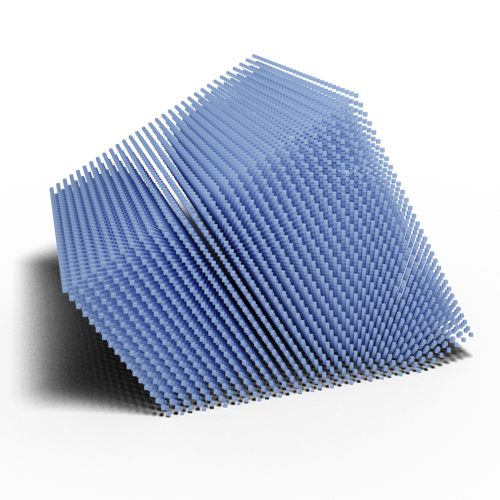} &
        \visualhullimage{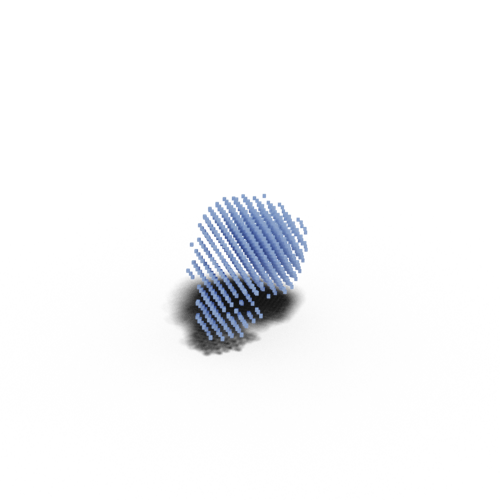} &
        \visualhullimage{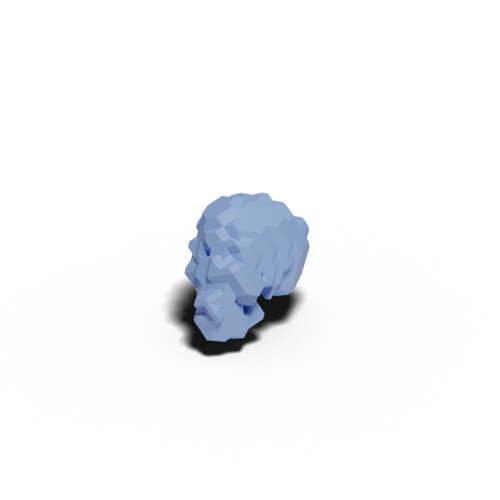} \\
        \footnotesize{Grid} & \footnotesize{Remaining points} & \footnotesize{Marching cubes mesh}  \\
    \end{tabular}
    \caption{Construction of the initial visual hull mesh for 3D reconstruction.}
    \label{fig:visual_hull_sup}
\end{figure}
}

\newcommand{\ablationweightsimage}[1]{\raisebox{-0.5\height}{\includegraphics[width=0.39\columnwidth, trim=100 100 115 90]{images/ablation_weights/#1}}}

\newcommand{\figAblationWeights}{
\begin{figure*}[tb]
    \centering
    \setlength{\tabcolsep}{3pt}
    \begin{tabular}{@{}ccccc@{}}
        \ablationweightsimage{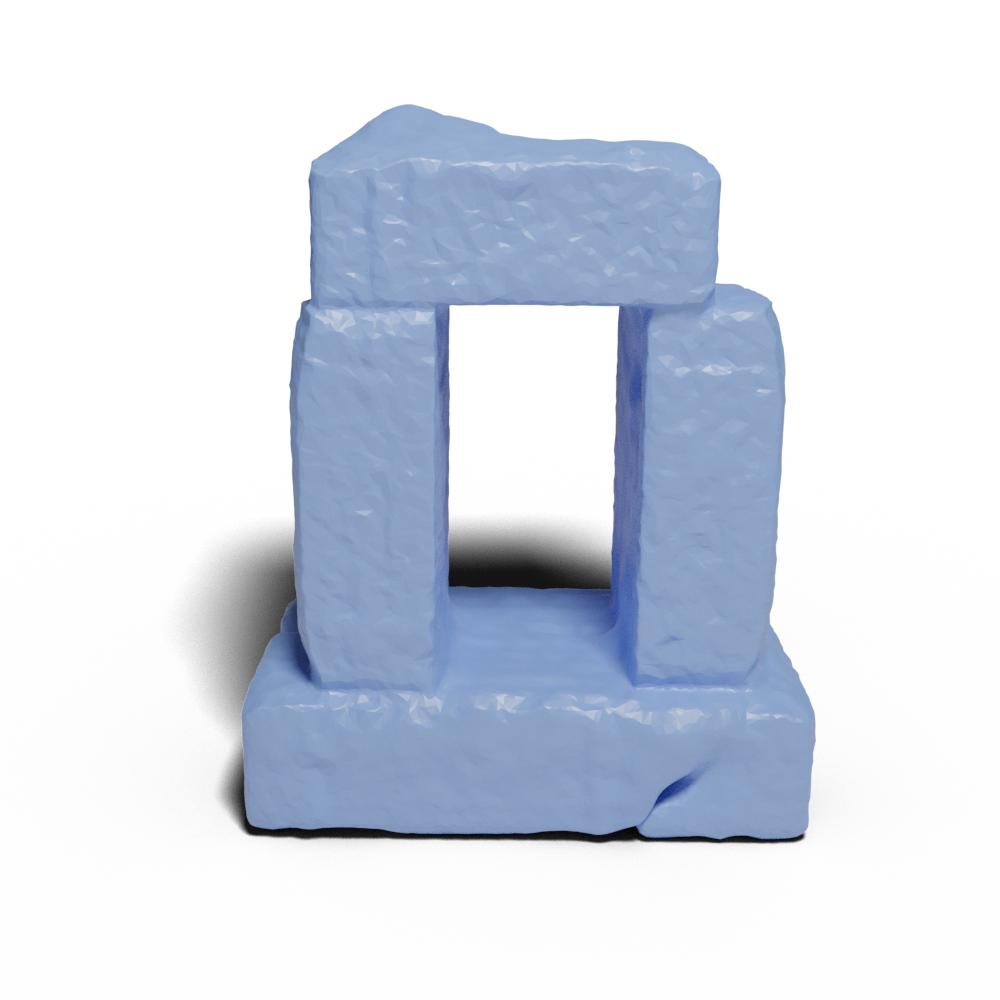} &
        \ablationweightsimage{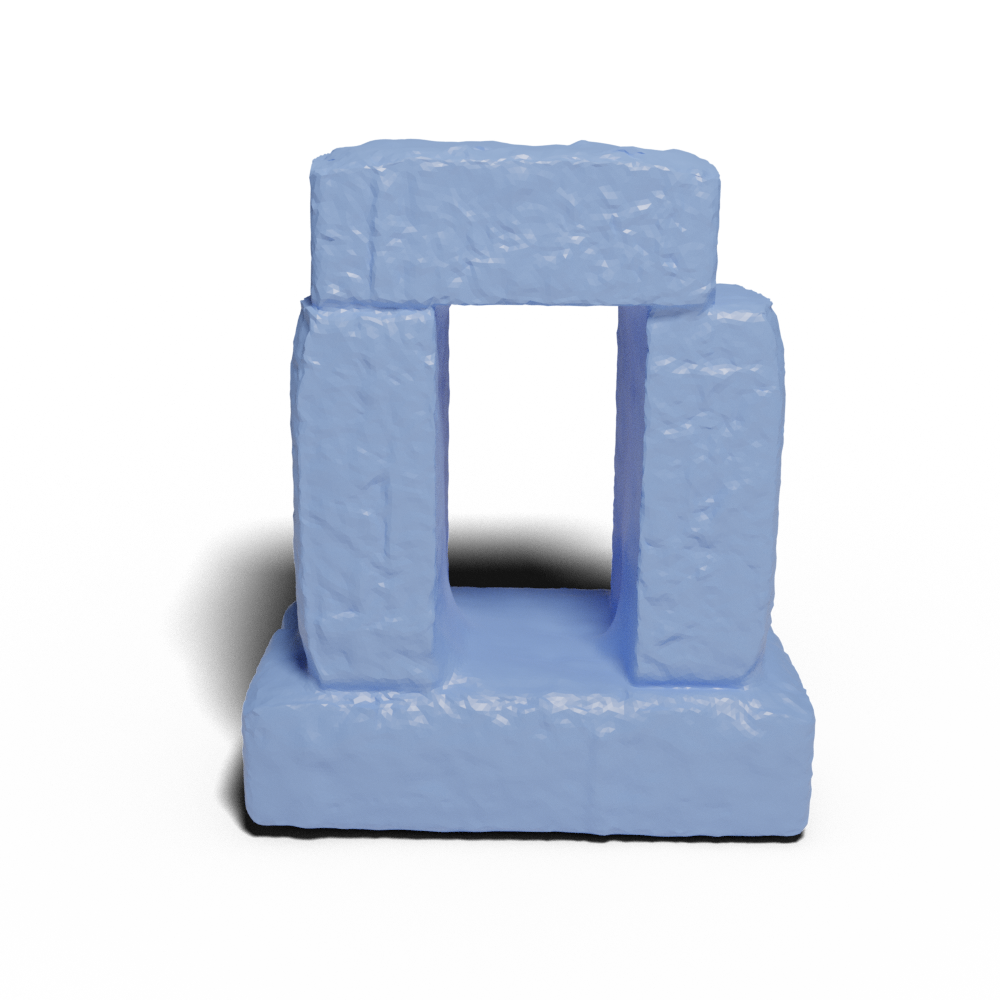} &
        \ablationweightsimage{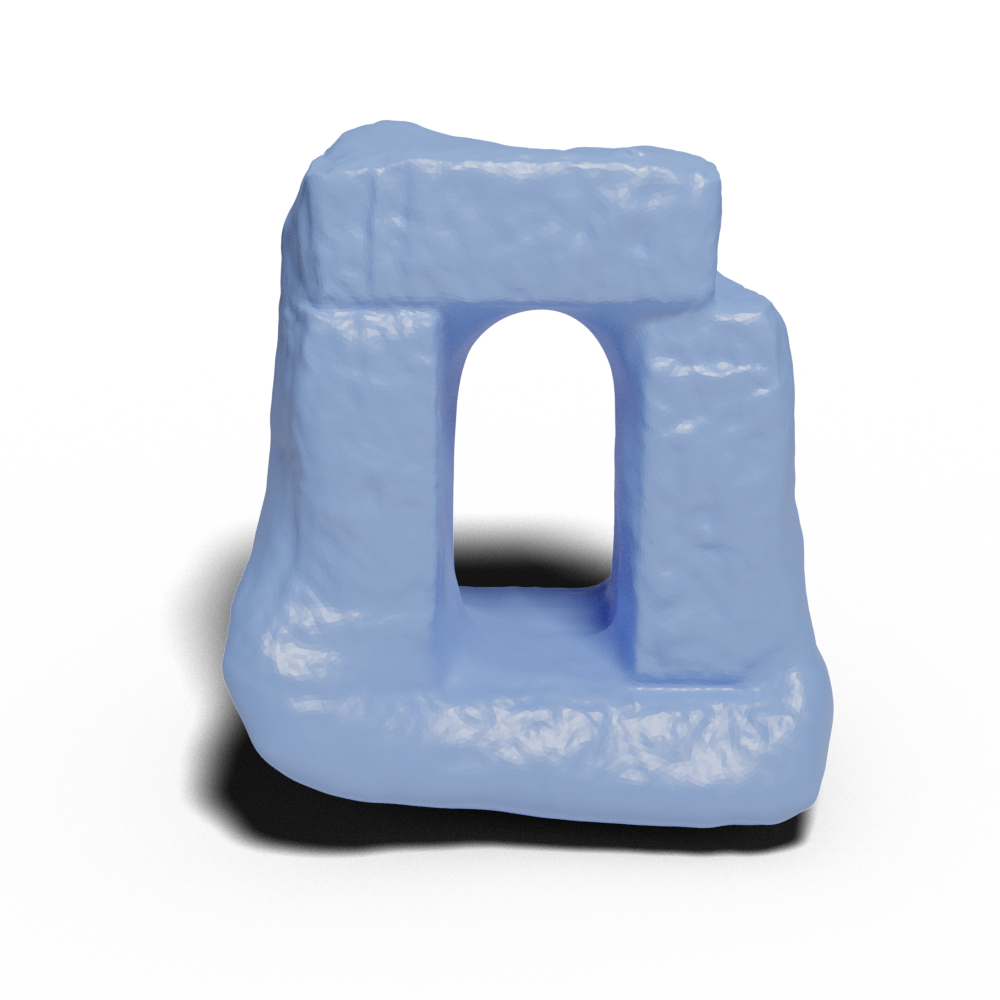} &
        \ablationweightsimage{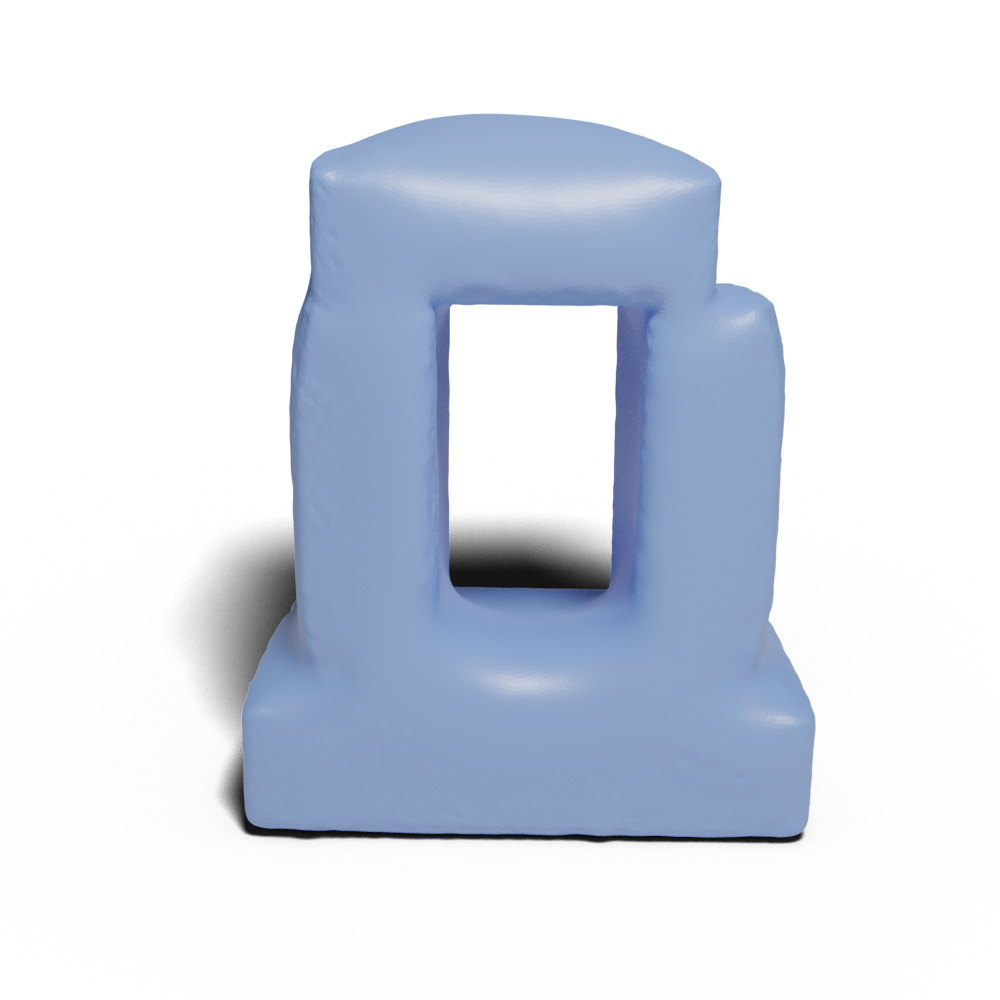} &
        \ablationweightsimage{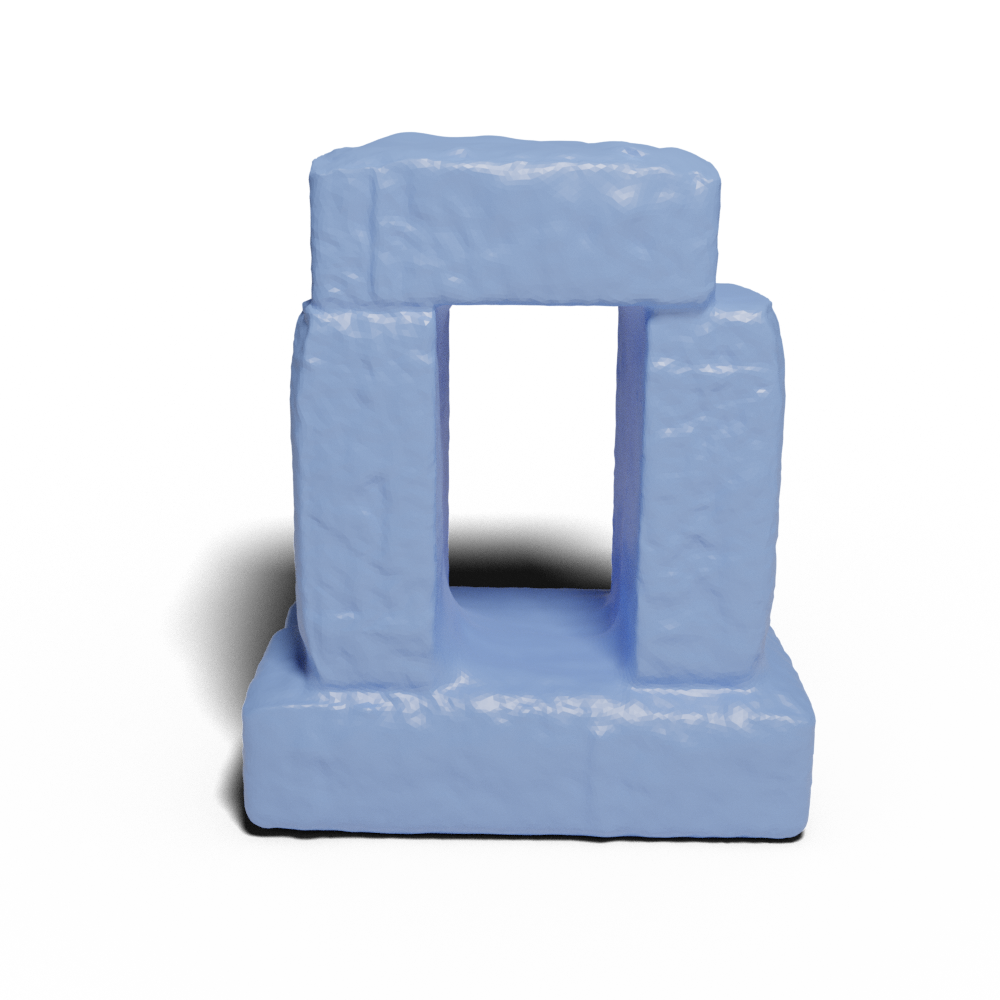}
        \\
        \ablationweightsimage{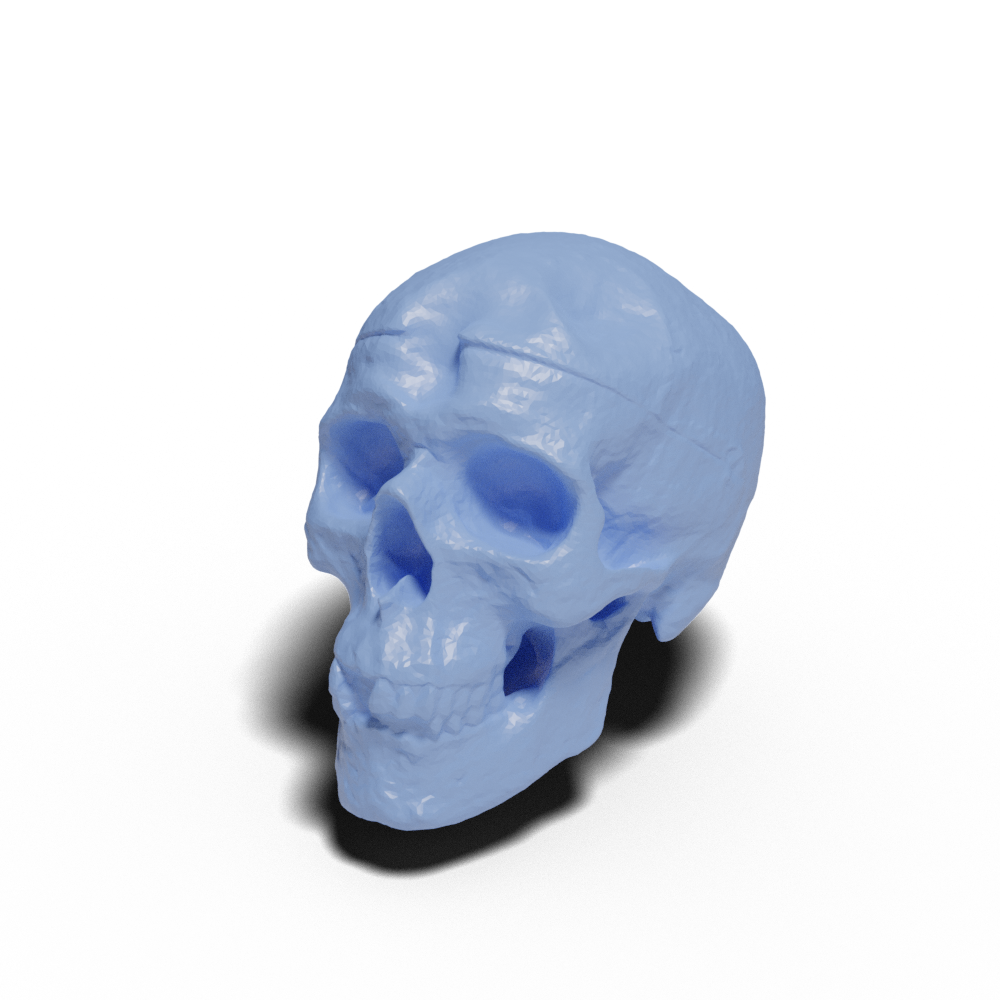} &
        \ablationweightsimage{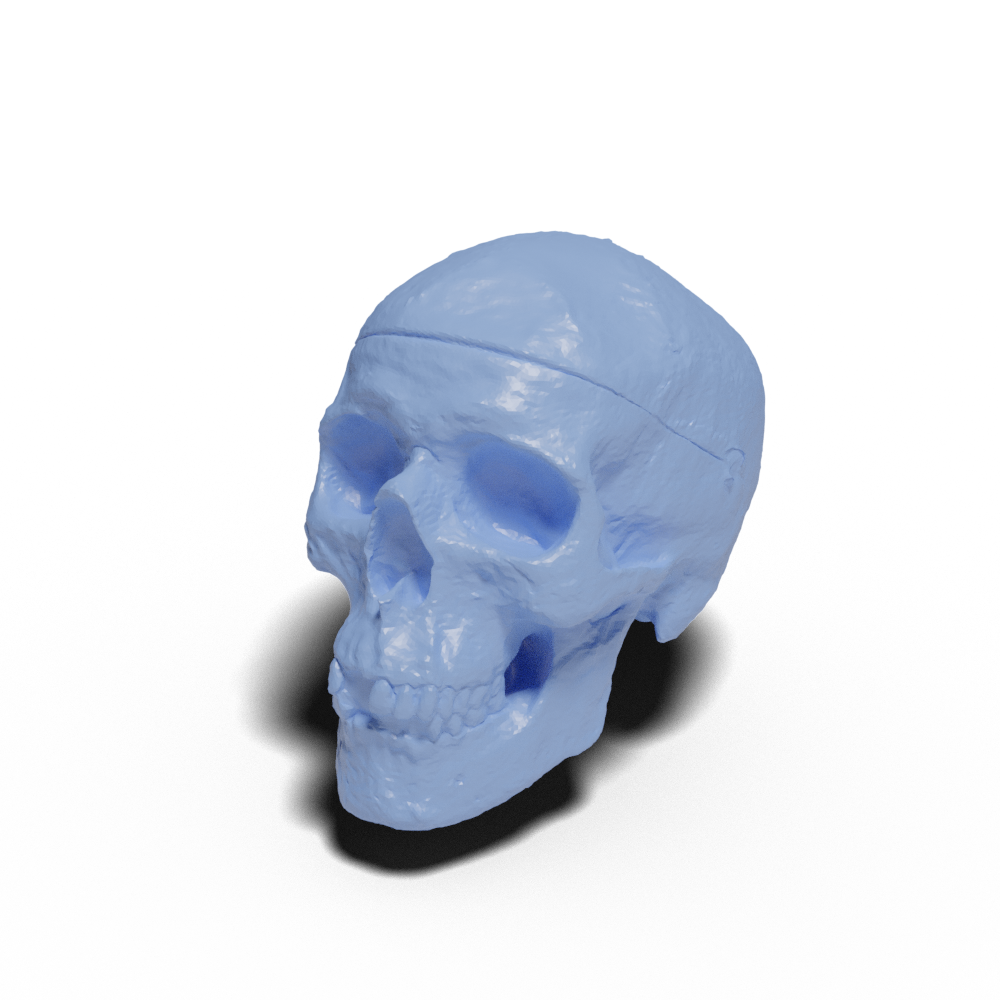} &
        \ablationweightsimage{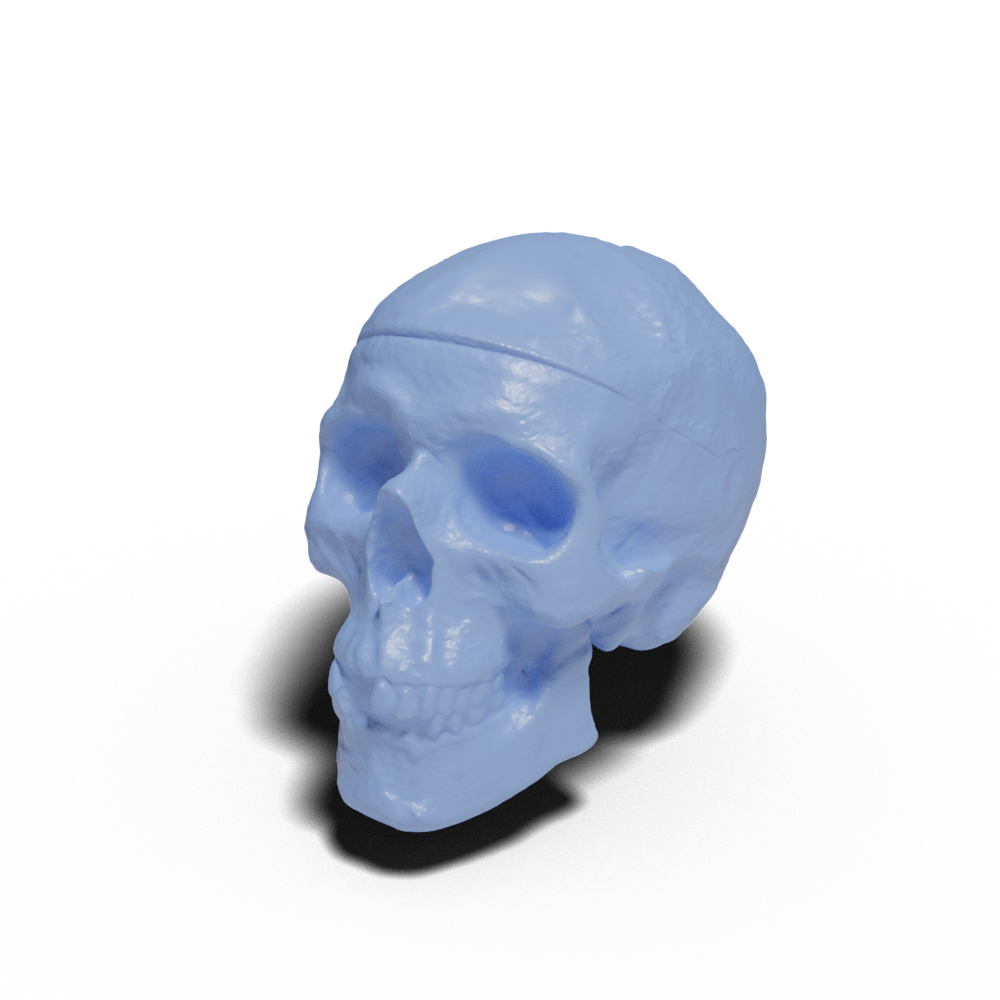} &
        \ablationweightsimage{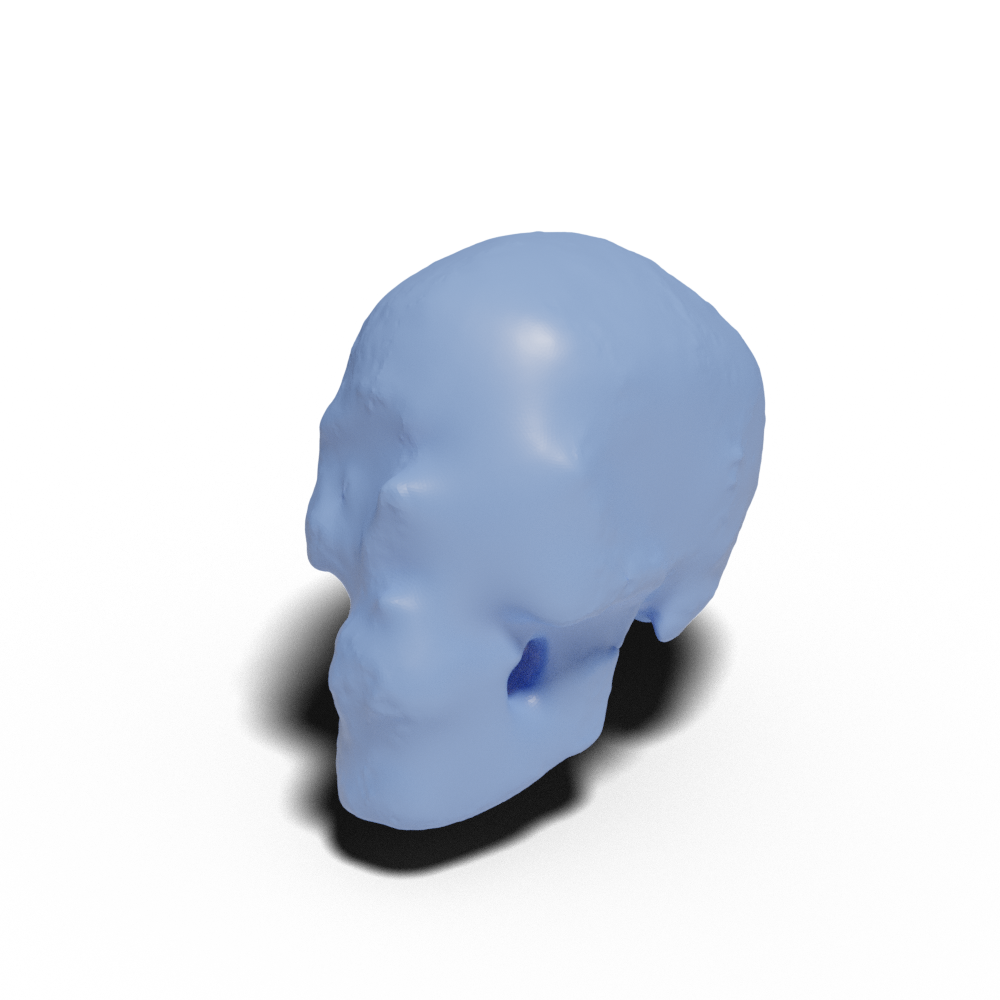} &
        \ablationweightsimage{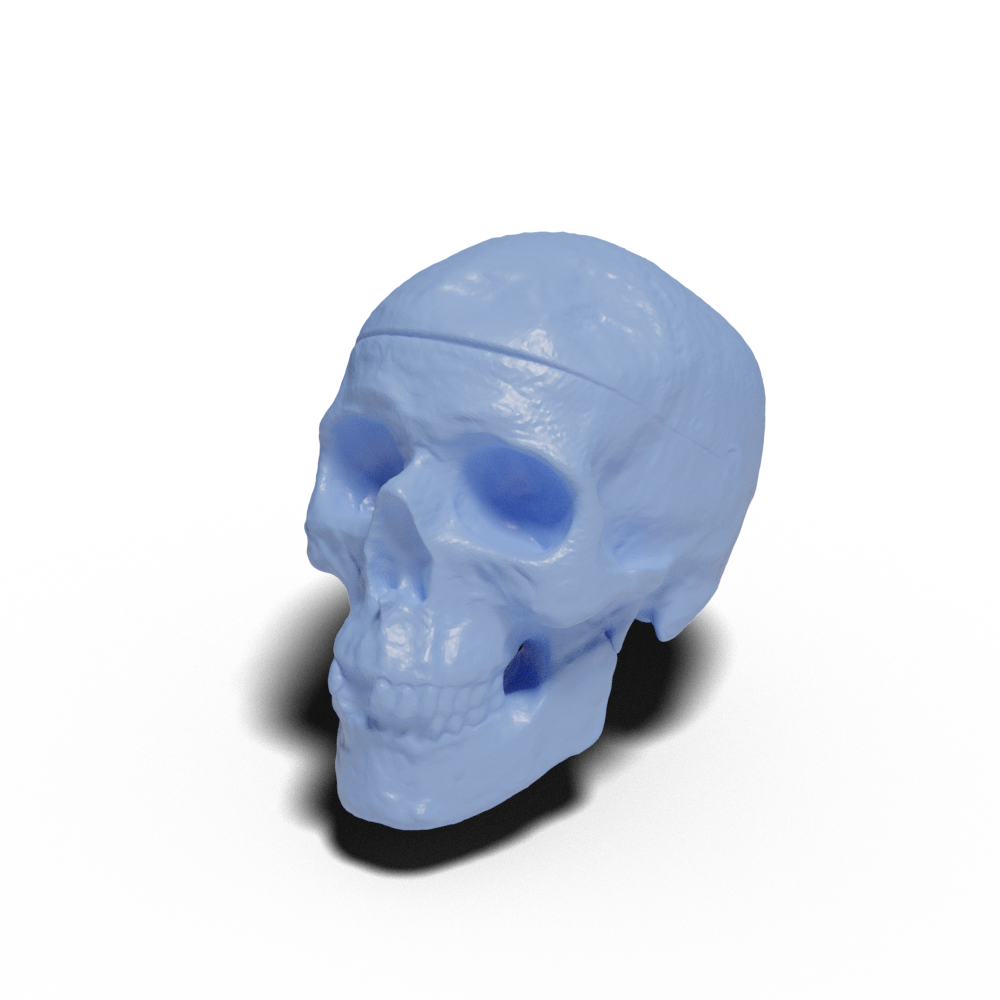}
        \\
        \small{No Laplacian term} & \small{No normal term} & \small{No silhouette term} &  \small{No shading term} & \small{Ours}
    \end{tabular}
    \caption{Ablation study of the objective function.}
    \label{fig:ablation_objective_function}
\end{figure*}
}

\usepackage[pagebackref,breaklinks,colorlinks]{hyperref}

\usepackage[capitalize]{cleveref}
\crefname{section}{Sec.}{Secs.}
\Crefname{section}{Section}{Sections}
\Crefname{table}{Table}{Tables}
\crefname{table}{Tab.}{Tabs.}

\def\cvprPaperID{5862} %
\def\confName{CVPR}
\def\confYear{2022}

\begin{document}

\title{\papertitle{}}

\author{\paperauthors{}}

\maketitle

\begin{abstract}
    We propose an analysis-by-synthesis method for fast multi-view 3D reconstruction of opaque objects with arbitrary materials and illumination. State-of-the-art methods use both neural surface representations and neural rendering. While flexible, neural surface representations are a significant bottleneck in optimization runtime. Instead, we represent surfaces as triangle meshes and build a differentiable rendering pipeline around triangle rasterization and neural shading. The renderer is used in a gradient descent optimization where both a triangle mesh and a neural shader are jointly optimized to reproduce the multi-view images. We evaluate our method on a public 3D reconstruction dataset and show that it can match the reconstruction accuracy of traditional baselines and neural approaches while surpassing them in optimization runtime. Additionally, we investigate the shader and find that it learns an interpretable representation of appearance, enabling applications such as 3D material editing.
\end{abstract}

\section{Introduction}

The reconstruction of 3D objects based on multiple images is a long standing problem in computer vision. Traditionally, it has been approached by matching pixels between images, often based on photo-consistency constraints or learned features~\cite{Han:2021:ImageBased3DObjectReconstruction, Laga:2020:SurveyDeepStereo}.
More recently, \emph{analysis-by-synthesis}%
, a technique built around the rendering operation, has re-emerged as a promising direction for reconstructing scenes with complex illumination, materials and geometry~\cite{Lyu:2020:DiffRefrac, Luan:2021:Unified, Mildenhall:2020:NeRF,Lombardi:2019:NeuralVolumes,Yu:2020:PixelNerf, Niemeyer:2020:DVR}.
At its core, parameters of a virtual scene are optimized so that its \emph{rendered appearance} from the input camera views matches the camera images. If the reconstruction focuses on solid objects, these parameters usually include a representation of the object surface.

\figTeaser

In gradient descent-based optimizations, analysis-by-synthesis for surfaces is approached differently depending on the differentiable rendering operation at hand. Methods that physically model light transport typically build on prior information such as light and material models~\cite{Loubet:2019:RDI, Luan:2021:Unified}. It is common to represent object surfaces with triangle meshes and use differentiable path tracers
(\eg{},~\cite{NimierDavid:2019:Mitsuba2, Li:2018:Redner, Zhang:2020:PathDiffRend}) to jointly optimize the geometry and parameters like the light position or material diffuse albedo. Due to the inherent priors, these methods do not generalize to arbitrary scenes.

Other methods instead model the rendering operation with neural networks~\cite{Niemeyer:2020:DVR, Yariv:2020:multiview, Oechsle:2021:UNISURF}, \ie, the interaction of material, geometry and light is partially or fully encoded in the network weights, without any explicit priors. Surfaces are often represented with implicit functions or more specifically \emph{implicit neural representations}~\cite{Park:2019:DeepSDF, Liu:2019:LearningImplicit, Niemeyer:2020:DVR} where the indicator function is modeled by a multi-layer perceptron (MLP) or any other form of neural network and optimized with the rendering networks in an end-to-end fashion.

While fully neural approaches are general, both in terms of geometry and appearance, current methods exhibit excessive runtime, making them impractical for domains that handle a large number of objects or multi-view video (\eg{} of human performances~\cite{Starck:2007:SurfaceCapturePerformance, Vlasic:2009:DynamicShapeCapture, Guo:2019:Relightables, Collet:2015:HQStreamVideo, Schreer:2019:CPVV}).

We propose Neural Deferred Shading (NDS), a fast analysis-by-synthesis method that combines triangle meshes and neural rendering. The rendering pipeline is inspired by real-time graphics and implements a technique called \emph{deferred shading}~\cite{Deering:1988:DeferredShading}: a triangle mesh is first rasterized and the pixels are then processed by a neural shader that models the interaction of geometry, material, and light. Since the rendering pipeline, including rasterization and shading, is differentiable, we can optimize the neural shader and the surface mesh with gradient descent (Figure~\ref{fig:teaser}). The explicit geometry representation enables fast convergence while the neural shader maintains the generality of the modeled appearance. Since triangle meshes are ubiquitously supported, our method can also be readily integrated with existing reconstruction and graphics pipelines.
Our technical contributions include:
\begin{itemize}
    \item A fast analysis-by-synthesis pipeline based on triangle meshes and neural shading that handles arbitrary illumination and materials
    \item A runtime decomposition of our method and a state-of-the-art neural approach
    \item An analysis of the neural shader and the influence of its parameters
\end{itemize}

\figOverview
\section{Related Work}

\subsection{Multi-View Mesh Reconstruction} 

There is a vast body of work on image-based 3D reconstruction for different geometry representations (\eg{} voxel grids, point clouds and triangle meshes). Here, we will only focus on methods that output meshes and refer to Seitz \etal{}~\cite{Seitz:2006:Survey} for an overview of other approaches.

\paragraph{Photo-Consistency.}

During the past decades, multi-view methods have primarily exploited photo-consistency across images.
Most of these approaches traverse different geometry representations like depth maps or point clouds before extracting (and further refining) a mesh, \eg{} \cite{Campbell:2008:DepthMaps, Hiep:2009:HighResLargeScaleStereo, Furukawa:2010:Accurate, Tola:2012:efficient, Galliani:2015:Massively, Vu:2012:HighAccMultiView, BlumenthalBarby:2014:HiResDepth, Tamura:2020:TD3M}.
Some methods directly estimate a mesh by deforming or carving an initial mesh (\eg{} the visual hull) while minimizing an energy based on cross-image agreement~\cite{Fua:1995:ObjectCenteredSurfaceReconstruction, Zhang:2000:ImageBasedMultiRes, Isidro:2003:StochRefinement, Esteban:2003:SilhouetteStereoFusion, Furukawa:2006:CarvedVisualHull}.
Recently, learned image features and neural shape priors have been used to drive the mesh deformation process~\cite{Wen:2019, Lin:2019:PhotometricMeshOptim}.
Our method is similar to full mesh-based approaches in the sense that we do not use intermediate geometry representations. However, we also do not impose strict assumptions on object appearance across images, which enables us to handle non-Lambertian surfaces and varying light conditions.

\paragraph{Analysis-by-Synthesis.}

More than 20 years ago, Rockwood and Winget~\cite{Rockwood:1997:ObjRecFromImages} proposed to deform a mesh so that \emph{synthesized} images match the input images. Their early analysis-by-synthesis method builds on an objective function with similar terms as ours (and many modern approaches): shading, silhouette, and geometry regularization. Later works propose similar techniques (\eg{}~\cite{Yu:2004:ShapeAndRefl, Delaunoy:2008:Minimizing, Wu:2011:HighQualShape})%
, yet all either assume known material or light parameters or restrict the parameter space with prior information, \eg{} by assuming constant material across surfaces. In contrast, we optimize all parameters of the virtual scene and do not assume specific material or light models.

Optimizing many parameters of a complex scene, including geometry, material and light, has only lately become practical, arguably with the advent of differentiable rendering. Differentiable path tracers have been used on top of triangle meshes to recover not only the geometry but also the (spatially varying) reflectance and light~\cite{Loubet:2019:RDI, Luan:2021:Unified}, only from images. Related techniques can reconstruct transparent objects~\cite{Lyu:2020:DiffRefrac}. Similarly, we perform analysis-by-synthesis by optimizing a mesh with differentiable rendering. However, we use rasterization and do not simulate light transport. In our framework, the view-dependent appearance is learned by a neural shader, which neither depends on material or light models nor imposes constraints on the acquisition setup (\eg{} co-located camera and light).

Besides mesh reconstruction from real world images, analysis-by-synthesis with differentiable rendering has recently been used for image-based geometry processing~\cite{Liu:2018:Paparazzi} and appearance-driven mesh simplification~\cite{Hasselgren:2021}. Similar to us, these approaches deform a triangle mesh to reproduce target images, albeit their targets are fully synthetic.

\subsection{Neural Rendering and Reconstruction}

In this work, we understand neural rendering as training and using a neural network to synthesize color images from 2D input (\eg{} semantic labels or UV coordinates), recently named ``2D Neural Rendering"~\cite{Tewari:2021:AdvancesNeuralRendering}.
Neural rendering has been used as integral part of 3D reconstruction methods with neural scene representations.

Introduced by Mildenhall \etal{}~\cite{Mildenhall:2020:NeRF}, neural radiance fields are volumetric scene representations used for 3D reconstruction, which are trained to output RGB values and volume densities at points along rays casted from different views. This idea has been adapted by a large number of recent works~\cite{Dellaert:2021:NerfSurvey}. Although not strictly based on neural rendering, these methods are related to ours by their analysis-by-synthesis characteristics. While the volumetric representation can handle transparent objects, most methods focus on view synthesis, so extracted surfaces lack geometric accuracy. Lassner~\etal{}~\cite{Lassner:2020:Pulsar} propose a volumetric representation based on translucent spheres, which are shaded with neural rendering. Similar to us, they jointly optimize geometry and appearance with a focus on speed, yet most details in their reconstruction are not present in the geometry but ``hallucinated" by the neural renderer.

Implicit surfaces encoded in neural networks are another popular geometry representation for 3D reconstruction, most notably occupancy networks~\cite{Mescheder:2019:OccupancyNetworks, Niemeyer:2020:DVR, Oechsle:2021:UNISURF, Peng:2020:Convolutional} and neural signed distance functions~\cite{Park:2019:DeepSDF,Yariv:2020:multiview,Kellnhofer:2021:NLR,Wang:2021:Neus}.
Here, surfaces are implicitly defined by level sets. For 3D reconstruction, these geometry networks are commonly trained end-to-end with a neural renderer to synthesize scenes that reproduce the input images. 
We also use neural rendering to model the appearance, but represent geometry explicitly with triangle meshes, which can be efficiently optimized and readily integrated into existing graphics workflows.

Similar to us, Thies~\etal{}~\cite{Thies:2019:NeuralTextures} present a deferred mesh renderer with neural shading. However, their convolutional neural network-based renderer can ``hallunicate" colors at image regions that are not covered by geometry, as opposed to our MLP-based shader. Most notably, their method aims at view synthesis, therefore only the renderer weights are optimized, while the mesh vertices remain unchanged.

\section{Method}

Given a set of images $\images = \{\image_1, \cdots, \image_{n}\}$
from calibrated cameras and corresponding masks $\masks = \{\mask_1, \cdots, \mask_{n}\}$, 
we want to estimate the 3D surface of an object shown in the images. To this end, we follow an analysis-by-synthesis approach: we find a surface that reproduces the images when rendered from the camera views. In this work, the surface is represented by a triangle mesh $\mesh = (\vertices, \edges, \faces)$, consisting of vertex positions \vertices, a set of edges \edges, and a set of faces \faces. We solve the optimization problem using gradient descent and gradually deform a mesh based on an objective function that compares renderings of the mesh to the input images. 

Faithfully reproducing the images via rendering requires an estimate of the surface material and illumination if we simulate light transport, \eg with a differentiable path tracer~\cite{Li:2018:Redner, NimierDavid:2019:Mitsuba2}. However, because our focus is mainly on the geometry, we do not accurately estimate these quantities and thus also avoid the limitations imposed by material and light models. Instead, we propose a differentiable mesh renderer that implements a deferred shading pipeline and handles arbitrary materials and light settings. At its core, a differentiable rasterizer produces geometry maps per view, which are then processed by a learned shader. See Figure~\ref{fig:pipeline} for an overview.

\subsection{Neural Deferred Shading}

\figShaderArchitecture

Our differentiable mesh renderer follows the structure of a deferred shading pipeline from real-time graphics: Given a camera $i$, the mesh is rasterized in a first pass, yielding a triangle index and barycentric coordinates \emph{per pixel}. This information is used to interpolate both vertex positions and vertex normals, creating a geometry buffer (g-buffer) with per-pixel positions and normals. In a second pass, the g-buffer is processed by a learned shader
\begin{equation}
f_\theta(\mathbf{x}, \mathbf{n}, \boldsymbol{\omega}_o) \in [0, 1]^3
\end{equation}
with parameters $\theta$. The shader returns an RGB color value for a given position $\mathbf{x} \in \mathbb{R}^3$, normal $\mathbf{n} \in \mathbb{R}^3$, and view direction $\boldsymbol{\omega}_o = \frac{\mathbf{c}_i - \mathbf{x}}{\Vert \mathbf{c}_i - \mathbf{x} \Vert}$, with $\mathbf{c}_i \in \mathbb{R}^3$ the center of camera~$i$. It encapsulates the appearance, \ie, interaction of geometry, material and light as well as the camera pixel response, and is optimized together with the geometry. We represent the shader as a shallow multi-layer perceptron, with $\theta$ as the parameters of the fully-connected layers (Figure~\ref{fig:shader_architecture}). In this context, it has been shown that providing the normal and view direction with the position is necessary for disentangling the geometry from the appearance~\cite{Yariv:2020:multiview}.

In addition to a color image, the renderer also produces a mask that indicates if a pixel is covered by the mesh.

\subsection{Objective Function}

Finding an estimate of shape and appearance formally corresponds to solving the following minimization problem in our framework
\begin{equation}
    \argmin_{\vertices, \theta} L_\text{appearance}(\mesh, \theta; \images, \masks) + L_\text{geometry}(\mesh),
\end{equation}
where $L_\text{appearance}$ compares the rendered appearance of the estimated surface to the camera images and $L_\text{geometry}$ regularizes the mesh to avoid undesired vertex configurations.

\subsubsection{Appearance}

The appearance objective is composed of two terms
\begin{equation}
    L_\text{appearance} = L_\text{shading} +  L_\text{silhouette},
\end{equation}
where the shading term
\begin{equation}
    L_\text{shading} = \lambda_{\text{shading}} \frac{1}{\setcard{\images}} \sum_{i = 1}^{\setcard{\images}} \Vert \image_i - \tilde{\image}_i \Vert_1
\end{equation}
ensures that the color images produced by the shader $\tilde{\image}_i$ correspond to the input images and the silhouette term
\begin{equation}
    L_\text{silhouette} = \lambda_{\text{silhouette}} \frac{1}{\setcard{\masks}} \sum_{i = 1}^{\setcard{\masks}} \Vert \mask_i - \tilde{\mask}_i \Vert_1
\end{equation}
ensures that the rendered masks $\tilde{\mask}_i$ match the input masks for all views. Here, $\Vert \cdot \Vert_1$ denotes the mean absolute error of all pixels in an image. Formally, the masks $\tilde{\mask}_i$ are functions of the geometry $\mesh{}$ and the parameters of camera $i$, while the color images $\tilde{\image}_i$ are also functions of the neural shader (or more precisely its parameters $\theta$).

Separating the shading from the silhouette objective mainly has performance reasons: For a camera view $i$, the rasterization considers all pixels in the image, therefore computing the mask $\tilde{\mask}_i$ is cheap. However, shading is more involved and requires invoking the neural shader for all pixels after rasterization, which is an expensive operation. In practice, we only shade a subset of pixels inside the intersection of input and rendered masks while comparing the silhouette for all pixels. Additionally, we also limit the number of camera views considered in each gradient descent iteration.

\figReconstructionDTU

\subsubsection{Geometry Regularization}

Naively moving the vertices unconstrained in each iteration quickly leads to undesirable meshes with degenerate triangles and self-intersections.
We use a geometry regularization term that favors smooth solutions and is inspired by Luan \etal{}~\cite{Luan:2021:Unified}:
\begin{equation}
    L_\text{geometry} = L_\text{laplacian} + L_\text{normal}.
\end{equation}
Let $\vertices \in \mathbb{R}^{n \times 3}$ be a matrix with vertex positions as rows, the Laplacian term is defined as
\begin{equation}
L_\text{laplacian} = \lambda_\text{laplacian} \frac{1}{n} \sum_{i=1}^{n} \Vert \boldsymbol{\delta}_i \Vert_2^2,
\end{equation}
where 
\begin{equation}
\boldsymbol{\delta}_i = (\laplacian{} V)_i \in \mathbb{R}^3
\end{equation}
are the differential coordinates~\cite{Alexa:2003:Differential} 
of vertex~$i$, $\laplacian{} \in \mathbb{R}^{n \times n}$ is the graph Laplacian of the mesh \mesh{} and $\Vert \cdot \Vert_2$ is the Euclidean norm. Intuitively, by minimizing the magnitude of the differential coordinates of a vertex, we minimize its distance to the average position of its neighbors.

The normal consistency term is defined as
\begin{equation}
    L_\text{normal} = \lambda_\text{normal} \frac{1}{\vert \bar{\faces} \vert} \sum_{(i, j) \in \bar{\faces}} (1 - \normal_i \cdot \normal_j)^2,
\end{equation}
where $\bar{\faces}$ is the set of triangle pairs that share an edge and $\normal_i \in \mathbb{R}^3$ is the normal of triangle $i$ (under an arbitrary ordering of the triangles). It computes the cosine similarity between neighboring face normals and enforces additional smoothness.

While some prior work (\eg, \cite{Liu:2018:Paparazzi, Luan:2021:Unified}) uses El Topo~\cite{Brochu:2009:ElTopo} for robust mesh evolution, we found that our geometric regularization sufficiently avoids degenerate vertex configurations. Without El Topo, we are unable to handle topology changes but avoid its impact on runtime performance. 

\subsection{Optimization}

Our optimization starts from an initial mesh that is computed from the masks and resembles a visual hull~\cite{Laurentini:1994:VisualHull}. Alternatively, it can start from a custom mesh.

Similar to prior work, we use a coarse-to-fine scheme for the geometry: Starting from a coarsely triangulated mesh, we progressively increase its resolution during optimization. Inspired by Nicolet \etal{}~\cite{Nicolet:2021:LargeSteps} we remesh the surface with the method of Botsch and Kobbelt~\cite{Botsch:2004:Remeshing}, halving the average edge length multiple times at fixed iterations. After mesh upsampling, we also increase the weights of the regularization terms by 4 and decrease the gradient descent step size for the vertices by 25\,\%, which we empirically found helps to improve the smoothness for highly tesselated meshes.

Since some quantities in geometry regularization (\eg{} graph Laplacian) only depend on the connectivity of the mesh, we save time by precomputing them once after upsampling and reusing them in the iterations after.

\section{Experimental Results}

We implemented our method on top of the automatic differentiation framework PyTorch~\cite{Paszke:2019:PyTorch} and use the \textsc{Adam}~\cite{Kingma:2015:Adam} optimizer for momentum-based gradient descent. Our differentiable rendering pipeline uses the high-performance primitives by Laine \etal{}~\cite{Laine:2020}. In our experiments, we run 2000 gradient descent iterations and remesh after 500, 1000, and 1500 iterations. We randomly select one view per iteration to compute the appearance term and shade 75\,\% of mask pixels. The individual objective terms are weighted with $\lambda_{\text{shading}} = 1$, $\lambda_{\text{silhouette}} = 2$, $\lambda_{\text{laplacian}} = 40$, and $\lambda_{\text{normal}}=0.1$. All time measurements were taken on a Windows workstation with an Intel Xeon 32$\times$2.1 GHz CPU, 128 GB of RAM, and an NVIDIA Titan RTX GPU with 24 GB of VRAM.

\subsection{3D Reconstruction}

We demonstrate that our method can be used for multi-view 3D reconstruction. Starting from a coarse visual hull-like mesh, it quickly converges to a reasonable estimate of the object surface. We test our method on the \textsc{DTU} multi-view dataset~\cite{Jensen:2014:DTU} with the object selection and masks from previous work~\cite{Niemeyer:2020:DVR, Yariv:2020:multiview}. We compare our results to two methods: (1) \textsc{Colmap}~\cite{Schoenberger:2016:sfm, Schoenberger:2016:mvs}, a traditional SfM pipeline that serves as a baseline, and (2) \textsc{IDR}~\cite{Yariv:2020:multiview}, a state-of-the-art analysis-by-synthesis method that uses neural signed distance functions as geometry representation. By default, our \colmap{} results include trimming (trim7) and we indicate untrimmed results explicitly (trim0).

\tabReconstructionDTU

Figure~\ref{fig:reconstruction_dtu} shows qualitative results for two objects from the DTU dataset and Table~\ref{table:reconstruction_dtu} contains quantitative results for all objects. We used the official DTU evaluation script to generate the Chamfer-L1 scores and benchmarked all workflows for the time measurements (including data loading times). For \idr{} and our method we disabled any intermediate visualizations.

In absolute terms (millimeters), the reconstruction accuracy of our method is close to both the traditional baseline and the state-of-the-art neural method. Although \colmap{} reconstructs many surfaces accurately, only IDR and our method properly handle regions dominated by view-dependent effects (\eg{} non-Lambertian materials) and produce watertight surfaces that can remain untrimmed. When \colmap{} is used without trimming, the reconstruction becomes more complete but it is less accurate than ours for some objects. Our method is limited by the topology and genus of the initial mesh and therefore cannot capture some geometric details that can be recovered with more flexible surface representations. We also observe that our surfaces are often not as smooth as \idr{}'s and concave regions are not as prominent. The latter is potentially related to the mesh stiffness induced by our geometry regularization.

On the other hand, our method is significantly faster: roughly 10 times faster than \colmap{} and 80 times faster than \idr{} in the default configuration. Since the number of iterations is a hyper parameter for \idr{} and our method (and also has different semantics), we show equal time results for a fair comparison of both (Figure~\ref{fig:reconstruction_dtu_equal_time}). Our method quickly converges to a detailed estimate, while \idr{} only recovers a rough shape in the same time. Even after 50 minutes, \idr{} still lacks details present in our result.

\figReconstructionDTUEqualTime

Since \idr{} and our method have similar architectures, \ie{}, both perform analysis-by-synthesis and use gradient descent to jointly optimize shape and appearance, we can meaningfully compare the runtime in more detail. Table~\ref{table:runtime} shows the runtime of \emph{one} gradient descent iteration (see the supplementary material for a full decomposition of our runtime).

An iteration in \idr{} takes roughly twice the time as in our method, with the majority of time spent for ray marching the implicit function. In contrast, the time taken for rasterizing the triangle mesh is negligible in our method. Most of the ray marching time in \idr{} can be attributed to evaluating the network, thus switching to a more shallow neural representation could be a way to reduce the runtime. The remaining time is spent on operations like root finding, which could be accelerated by a more optimized implementation.

However, the runtime difference in gradient descent iterations cannot be the sole reason for our fast convergence. Even though iterations in \idr{} require twice the time, it requires more than twice the total time to show the same level of detail as our method (see Figure~\ref{fig:reconstruction_dtu_equal_time}).

In our method, we noticed that after mesh upsampling finer details quickly appear in the geometry. Thus, our fast convergence time might partially be related to the fact that we can locally increase the geometric freedom with a finer tessellation, while \idr{} and similar methods have no \emph{explicit} control over geometric resolution.

\tabRuntime

\subsection{Mesh Refinement}

\figMeshRefinement

Many reconstruction workflows based on photo-consistency are very mature, established and deliver high-quality results. Yet, they can fail for challenging materials or different light conditions across images, producing smoothed outputs as a compromise to errors in photo-consistent matching.

Since our method can start from arbitrary triangle meshes, we propose the refinement of outputs from traditional pipelines as one possible application. Our method then acts as a \emph{post-processing step}, improving the given geometry or parts of it with global multi-view constraints.
We demonstrate this application on a human body dataset that contains \SI{360}{\degree} images of human subjects captured by 32 cameras and meshes from a traditional 3D reconstruction pipeline~\cite{Schreer:2019:CPVV}.

Figure~\ref{fig:mesh_refinement} shows refinement results for the heads of two persons. We use 1000 iterations and upsample the mesh once. Since the initial mesh is already a good estimate of the true surface but the neural shader is randomly initialized, we rebalance the gradient descent step sizes so that the shader progresses faster than the geometry. We are able to recover details that are lost in the initial reconstruction. Very fine details like the facial hair are still challenging and lead to slight noise in our refined meshes.

\subsection{Analysis of the Neural Shader}

\figShaderViewSynth

Although recovering geometry is the main focus of our work, investigating the neural shader after optimization can potentially give insights into the reconstruction procedure and the information that is encoded in the network.

Figure~\ref{fig:shader_view_synth} shows that rendering with a trained shader can be used for basic view synthesis, producing reasonable results for input and novel views. Thus, the shader seems to learn a meaningful representation of appearance, disentangled from the geometry. If desired, the view synthesis quality can be further improved by continuing the optimization of the shader while keeping the mesh fixed.

\figShaderPca

The neural shader is a composite of two functions (Figure~\ref{fig:shader_architecture})
\begin{equation}
    f_\theta(\mathbf{x}, \mathbf{n}, \boldsymbol{\omega}_o) = c\big(h(\mathbf{x}), \mathbf{n}, \boldsymbol{\omega}_o\big),
\end{equation}
where $h: \mathbb{R}^3 \to \mathbb{R}^{256}$ transforms points in 3D space to positional features and $c: (\mathbb{R}^{256} \times \mathbb{R}^3 \times \mathbb{R}^3) \to [0, 1]^3$ then extracts view-dependent colors. Both functions are dependent on parameters $\theta_h$ and $\theta_c$, respectively.
To further decompose the behavior of the shader, we perform a principle component analysis of the positional features from $h$ and project them to the two dominant components (Figure~\ref{fig:shader_pca}). The shader naturally learns similar positional features for regions with similar material, despite their distance in space. Variations in illumination also seem to be encoded in the positional features because shadowed regions have slightly different features than exposed regions of the same material. 

We investigate the behavior of the view-dependent part by replacing feature vectors in the positional latent space and then extracting colors with the function $c$. More specifically, we replace all features representing one material by a feature representing another material (Figure~\ref{fig:shader_pca}). The results suggest that the function $c$ reasonably generalizes to view and normal directions not encountered in combination with the replacement feature, which in this example leads to geometric features of the mesh being still perceivable and thus allows simple material editing.

\subsection{Ablation Studies}

\newcommand{\nsaimage}[1]{\raisebox{-0.5\height}{\includegraphics[width=0.39\columnwidth, trim=40 20 260 10, clip]{#1}}}

\begin{figure*}[htb]
    \centering
    \setlength{\tabcolsep}{4pt} %
    \setlength{\fboxsep}{0pt}
    \begin{tabular}{@{}ccccc@{}}
        \nsaimage{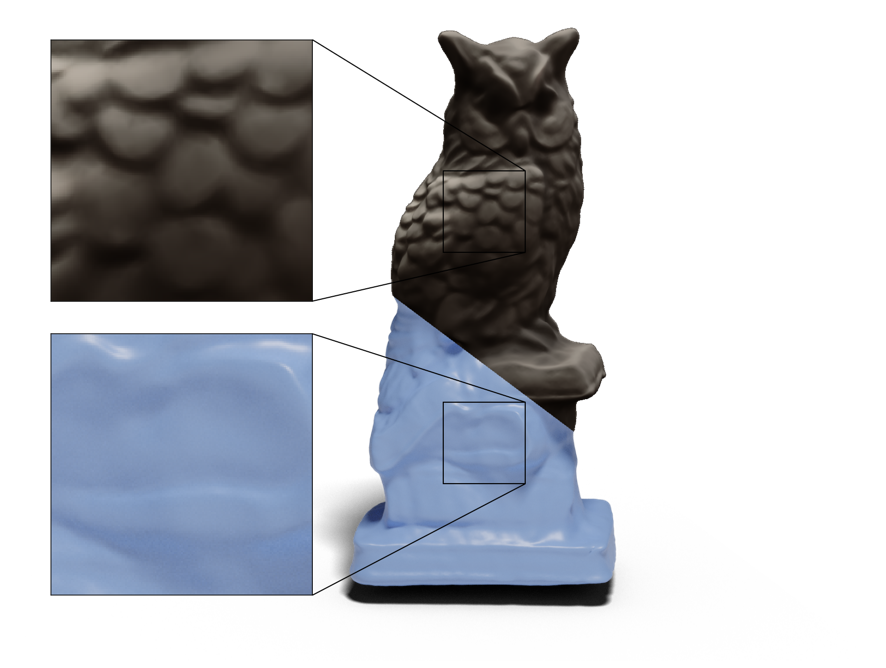} &
        \nsaimage{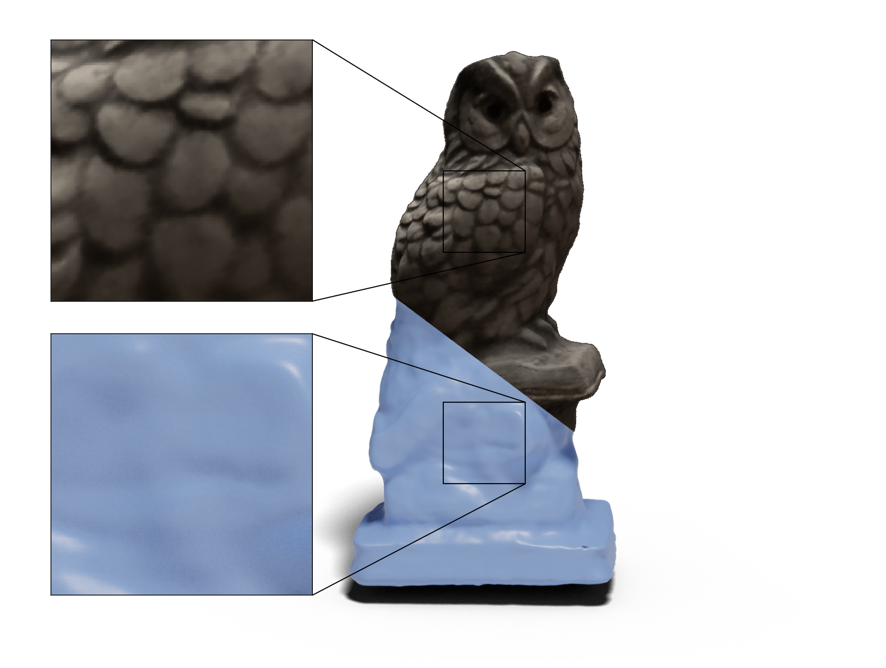} &
        \nsaimage{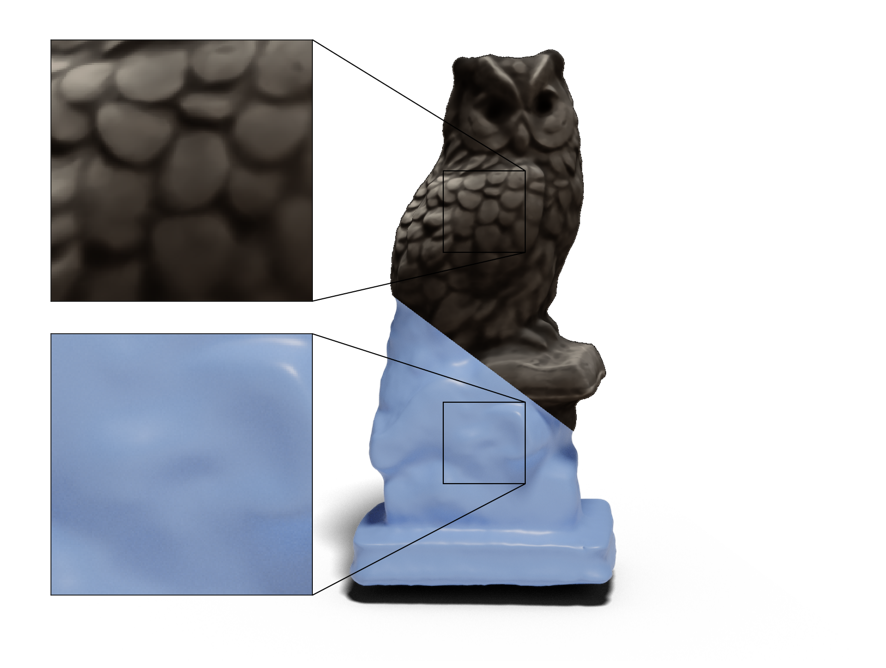} &
        \nsaimage{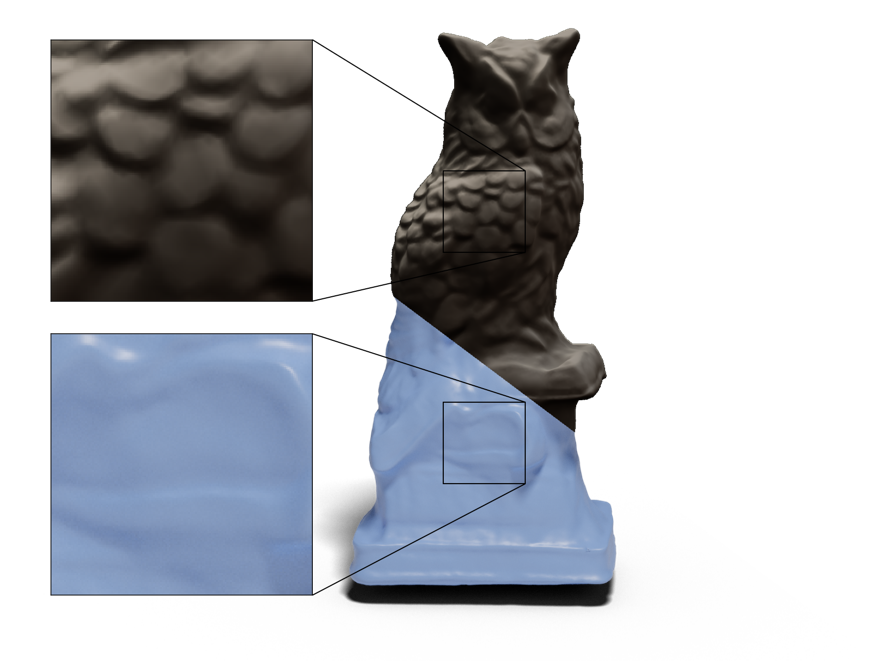} &
        \nsaimage{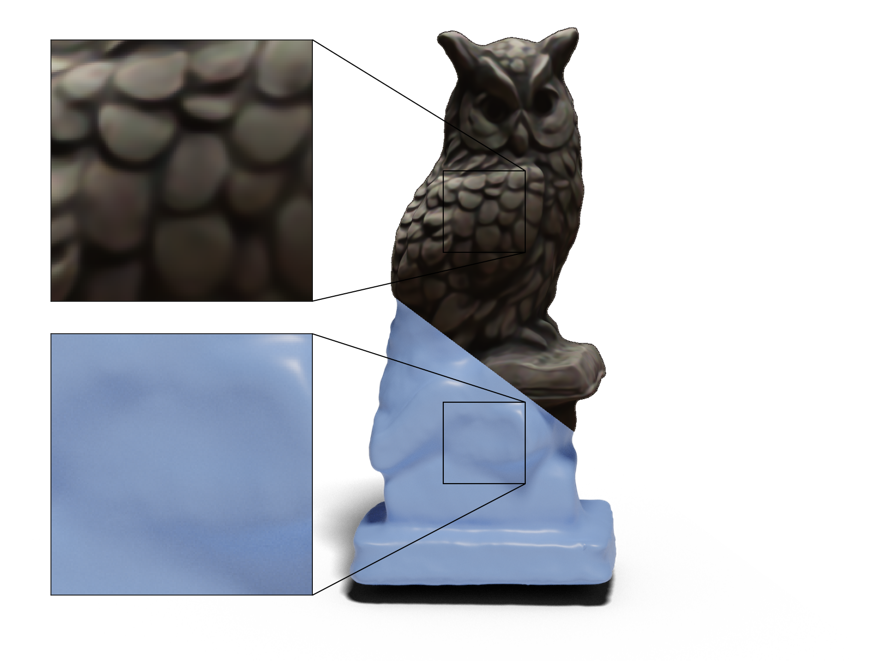} \\
        \small{PE4 (Ours)} & \small{PE10} & \small{GFF} & \small{ReLU} & \small{SIREN} \\
    \end{tabular}
    \caption{Ablation of different encodings for the positional features and activation functions. Note that the ears of the owl are only present in few views and not part of the reference geometry, thus are not considered in the comparison. For ReLU and SIREN we use no encoding.}
    \label{fig:neural_shader_ablation_positional_encoding}
\end{figure*}

We experiment with different encodings and network sizes for the position dependent part of the neural shader. In Figure~\ref{fig:neural_shader_ablation_positional_encoding}, we show the different results using positional encoding (PE)~\cite{Tancik:2020:FourierFeat}, Gaussian Fourier features (GFF)~\cite{Tancik:2020:FourierFeat}, sinusoidal activations (SIREN)~\cite{Sitzmann:2019:Siren} and a standard MLP with ReLU activations.

While some of these methods can generate acceptable renderings, they do not necessarily guarantee a sharper geometry. In particular, we observe that although the image rendered with SIREN or GFF has sharp features, the geometric detail of the mesh is less adequate. A possible explanation is that the network might quickly overfit and compensate for geometrical inaccuracies only in the appearance. Conversely, finding the correct direction along which to move the mesh vertices might be more difficult without positional encoding. In our experiments, we have obtained accurate reconstructions using positional encoding with 4 octaves, as opposed to the recommended 10~\cite{Mildenhall:2020:NeRF}.

We also examined the effect of different network sizes on the geometry and appearance (see supplementary). While there are no dramatic differences between the results, we observe that configurations with less than 2 layers or with more than 512 units per layer result in fewer geometric details. Additional ablation studies for the initial geometry and the objective function can also be found in the supplementary.

\section{Concluding Remarks}

We have presented a fast analysis-by-synthesis pipeline for 3D surface reconstruction from multiple images. Our method jointly optimizes a triangle mesh and a neural shader as representations of geometry and appearance to reproduce the input images, thereby combining the speed of triangle mesh optimization with the generality of neural rendering.

Our approach can match state-of-the-art methods in reconstruction accuracy, yet is significantly faster with average runtimes below 10 minutes. 

Using triangle meshes as geometry representation makes our method fully compatible with many traditional reconstruction workflows. Thus, instead of replacing the complete reconstruction architecture, our method can be integrated as a part, \eg{}, a refinement step.

Finally, a preliminary analysis of the neural shader suggests that it decomposes appearance in a natural way, which could help our understanding of neural rendering and enable simple ways for altering the scene appearance (\eg{} for material editing).

\paragraph{Limitations and Future Work.}

While triangle meshes are a simple and fast representation, using them robustly in 3D reconstruction with differentiable rendering is still challenging. We currently avoid undesired mesh configurations (\eg{} self-intersections) with carefully weighted geometry regularizers that steer the optimization towards smooth solutions. Finding an appropriate balance between smoothness and rendering terms is not always straightforward and can require fine-tuning for custom data.
In this regard, we are excited about integrating very recent work that proposes gradient preconditioning instead of geometry regularization~\cite{Nicolet:2021:LargeSteps}. 

Topology changes are a challenge for most mesh-based methods, including ours, and computationally expensive to handle (\eg{} with El Topo~\cite{Brochu:2009:ElTopo}). Likewise, adaptive reconstruction-aware subdivision would be preferable over standard remeshing, potentially including learned priors~\cite{Liu:2020:NeuralSubdivision}. Instead of moving vertices directly, a (partially pretrained) neural network could drive the deformation~\cite{Hanocka:2019:MeshCNN, Hanocka:2020:P2M}, making it more efficient, detail-aware, and less dependent on geometric regularization.

While neural shading is a powerful component of our system, allowing us to handle non-Lambertian surfaces and complex illumination, it is also a major obstruction for interpretability. The effect of changes to the network architecture can only be evaluated with careful experiments and often the black box shader behaves in non-intuitive ways. We experimentally provided preliminary insights but also think that a more thorough analysis is needed. Alternatively, physical light transport models could be combined with more specialized neural components (\eg{} irradiance or material) to isolate their effects. In this context, pre-training components to include learned priors also seems like a promising direction.

Although the shader can handle arbitrary material and light in theory, this claim needs more investigation. A possible path is exhaustive experiments with artificial scenes, starting with the simplest cases (how well does it handle a \emph{perfectly} Lambertian surface?).

\section*{Acknowledgements}

This work is part of the INVICTUS project that has received funding from the European Union’s Horizon 2020 research and innovation programme under grant agreement No 952147.

{\small
\bibliographystyle{ieee_fullname}
\bibliography{egbib}

\begin{thebibliography}{10}\itemsep=-1pt

\bibitem{Alexa:2003:Differential}
Marc Alexa.
\newblock Differential coordinates for local mesh morphing and deformation.
\newblock {\em The Visual Computer}, 19(2):105--114, 2003.

\bibitem{BlumenthalBarby:2014:HiResDepth}
David~C. Blumenthal-Barby and Peter Eisert.
\newblock High-resolution depth for binocular image-based modeling.
\newblock {\em Comput. Graph.}, 39:89–100, Apr. 2014.

\bibitem{Botsch:2004:Remeshing}
Mario Botsch and Leif Kobbelt.
\newblock A remeshing approach to multiresolution modeling.
\newblock In {\em Proceedings of the 2004 Eurographics/ACM SIGGRAPH Symposium
  on Geometry Processing}, SGP '04, page 185–192, New York, NY, USA, 2004.
  Association for Computing Machinery.

\bibitem{Bradski:2000:OpenCV}
G. Bradski.
\newblock {The OpenCV Library}.
\newblock {\em Dr. Dobb's Journal of Software Tools}, 2000.

\bibitem{Brochu:2009:ElTopo}
Tyson Brochu and Robert Bridson.
\newblock Robust topological operations for dynamic explicit surfaces.
\newblock {\em SIAM Journal on Scientific Computing}, 31(4):2472--2493, 2009.

\bibitem{Campbell:2008:DepthMaps}
Neill D.~F. Campbell, George Vogiatzis, Carlos Hern{\'a}ndez, and Roberto
  Cipolla.
\newblock Using multiple hypotheses to improve depth-maps for multi-view
  stereo.
\newblock In David Forsyth, Philip Torr, and Andrew Zisserman, editors, {\em
  Computer Vision -- ECCV 2008}, pages 766--779, Berlin, Heidelberg, 2008.
  Springer Berlin Heidelberg.

\bibitem{Clark:2021:Pillow}
Alex Clark et~al.
\newblock Pillow (pil fork).
\newblock \url{https://github.com/python-pillow/Pillow}, 2021.

\bibitem{Collet:2015:HQStreamVideo}
Alvaro Collet, Ming Chuang, Pat Sweeney, Don Gillett, Dennis Evseev, David
  Calabrese, Hugues Hoppe, Adam Kirk, and Steve Sullivan.
\newblock High-quality streamable free-viewpoint video.
\newblock {\em ACM Trans. Graph.}, 34(4), July 2015.

\bibitem{Dawson:2021:Trimesh}
Michael Dawson-Haggerty et~al.
\newblock Trimesh.
\newblock \url{https://github.com/mikedh/trimesh}.

\bibitem{Deering:1988:DeferredShading}
Michael Deering, Stephanie Winner, Bic Schediwy, Chris Duffy, and Neil Hunt.
\newblock The triangle processor and normal vector shader: A vlsi system for
  high performance graphics.
\newblock In {\em Proceedings of the 15th Annual Conference on Computer
  Graphics and Interactive Techniques}, SIGGRAPH '88, page 21–30, New York,
  NY, USA, 1988. Association for Computing Machinery.

\bibitem{Delaunoy:2008:Minimizing}
Amael Delaunoy, Emmanuel Prados, Pau Gargallo I~Pirac{\'e}s, Jean-Philippe
  Pons, and Peter Sturm.
\newblock {Minimizing the Multi-view Stereo Reprojection Error for Triangular
  Surface Meshes}.
\newblock In Mark Everingham, Chris~J. Needham, and Roberto Fraile, editors,
  {\em {BMVC 2008 - British Machine Vision Conference}}, pages 1--10, Leeds,
  United Kingdom, Sept. 2008. {BMVA}.

\bibitem{Dellaert:2021:NerfSurvey}
Frank Dellaert and Yen-Chen Lin.
\newblock Neural volume rendering: Nerf and beyond.
\newblock {\em CoRR}, abs/2101.05204, 2021.

\bibitem{Dombi:2020:ModernGL}
Szabolcs Dombi.
\newblock Moderngl, high performance python bindings for opengl 3.3+.
\newblock \url{https://github.com/moderngl/moderngl}.

\bibitem{DTU:2021:DTUMVS}
DTU.
\newblock {MVS} {D}ata {S}et -- 2014.
\newblock \url{https://roboimagedata.compute.dtu.dk/?page\_id=36}.

\bibitem{Esteban:2003:SilhouetteStereoFusion}
C.H. Esteban and F. Schmitt.
\newblock Silhouette and stereo fusion for 3d object modeling.
\newblock In {\em Fourth International Conference on 3-D Digital Imaging and
  Modeling, 2003. 3DIM 2003. Proceedings.}, pages 46--53, 2003.

\bibitem{Fua:1995:ObjectCenteredSurfaceReconstruction}
Pascal Fua and Yvan~G Leclerc.
\newblock Object-centered surface reconstruction: Combining multi-image stereo
  and shading.
\newblock {\em International Journal of Computer Vision}, 16(1):35--56, 1995.

\bibitem{Furukawa:2006:CarvedVisualHull}
Yasutaka Furukawa and Jean Ponce.
\newblock Carved visual hulls for image-based modeling.
\newblock volume~81, pages 564--577, 07 2006.

\bibitem{Furukawa:2010:Accurate}
Yasutaka Furukawa and Jean Ponce.
\newblock Accurate, dense, and robust multiview stereopsis.
\newblock {\em IEEE Transactions on Pattern Analysis and Machine Intelligence},
  32(8):1362--1376, 2010.

\bibitem{Galliani:2015:Massively}
Silvano Galliani, Katrin Lasinger, and Konrad Schindler.
\newblock Massively parallel multiview stereopsis by surface normal diffusion.
\newblock In {\em 2015 IEEE International Conference on Computer Vision
  (ICCV)}, pages 873--881, 2015.

\bibitem{Guo:2019:Relightables}
Kaiwen Guo, Peter Lincoln, Philip Davidson, Jay Busch, Xueming Yu, Matt Whalen,
  Geoff Harvey, Sergio Orts-Escolano, Rohit Pandey, Jason Dourgarian, Danhang
  Tang, Anastasia Tkach, Adarsh Kowdle, Emily Cooper, Mingsong Dou, Sean
  Fanello, Graham Fyffe, Christoph Rhemann, Jonathan Taylor, Paul Debevec, and
  Shahram Izadi.
\newblock The relightables: Volumetric performance capture of humans with
  realistic relighting.
\newblock {\em ACM Trans. Graph.}, 38(6), Nov. 2019.

\bibitem{Han:2021:ImageBased3DObjectReconstruction}
X. Han, H. Laga, and M. Bennamoun.
\newblock Image-based 3d object reconstruction: State-of-the-art and trends in
  the deep learning era.
\newblock {\em IEEE Transactions on Pattern Analysis and Machine Intelligence},
  43(05):1578--1604, may 2021.

\bibitem{Hanocka:2019:MeshCNN}
Rana Hanocka, Amir Hertz, Noa Fish, Raja Giryes, Shachar Fleishman, and Daniel
  Cohen-Or.
\newblock Meshcnn: A network with an edge.
\newblock {\em ACM Trans. Graph.}, 38(4), July 2019.

\bibitem{Hanocka:2020:P2M}
Rana Hanocka, Gal Metzer, Raja Giryes, and Daniel Cohen-Or.
\newblock Point2mesh: A self-prior for deformable meshes.
\newblock {\em ACM Trans. Graph.}, 39(4), July 2020.

\bibitem{Harris:2020:Numpy}
Charles~R. Harris, K.~Jarrod Millman, St{\'{e}}fan~J. van~der Walt, Ralf
  Gommers, Pauli Virtanen, David Cournapeau, Eric Wieser, Julian Taylor,
  Sebastian Berg, Nathaniel~J. Smith, Robert Kern, Matti Picus, Stephan Hoyer,
  Marten~H. van Kerkwijk, Matthew Brett, Allan Haldane, Jaime~Fern{\'{a}}ndez
  del R{\'{i}}o, Mark Wiebe, Pearu Peterson, Pierre G{\'{e}}rard-Marchant,
  Kevin Sheppard, Tyler Reddy, Warren Weckesser, Hameer Abbasi, Christoph
  Gohlke, and Travis~E. Oliphant.
\newblock Array programming with {NumPy}.
\newblock {\em Nature}, 585(7825):357--362, Sept. 2020.

\bibitem{Hasselgren:2021}
Jon Hasselgren, Jacob Munkberg, Jaakko Lehtinen, Miika Aittala, and Samuli
  Laine.
\newblock {Appearance-Driven Automatic 3D Model Simplification}.
\newblock In Adrien Bousseau and Morgan McGuire, editors, {\em Eurographics
  Symposium on Rendering - DL-only Track}. The Eurographics Association, 2021.

\bibitem{Hiep:2009:HighResLargeScaleStereo}
Vu~Hoang Hiep, Renaud Keriven, Patrick Labatut, and Jean-Philippe Pons.
\newblock Towards high-resolution large-scale multi-view stereo.
\newblock In {\em 2009 IEEE Conference on Computer Vision and Pattern
  Recognition}, pages 1430--1437, 2009.

\bibitem{Hunter:2007:Matplotlib}
J.~D. Hunter.
\newblock Matplotlib: A 2d graphics environment.
\newblock {\em Computing in Science \& Engineering}, 9(3):90--95, 2007.

\bibitem{Isidro:2003:StochRefinement}
Isidro and Sclaroff.
\newblock Stochastic refinement of the visual hull to satisfy photometric and
  silhouette consistency constraints.
\newblock In {\em Proceedings Ninth IEEE International Conference on Computer
  Vision}, pages 1335--1342 vol.2, 2003.

\bibitem{Jacobson:2018:igl}
Alec Jacobson, Daniele Panozzo, et~al.
\newblock {libigl}: A simple {C++} geometry processing library.
\newblock \url{https://libigl.github.io/}, 2018.

\bibitem{Jensen:2014:DTU}
Rasmus Jensen, Anders Dahl, George Vogiatzis, Engil Tola, and Henrik Aan{\ae}s.
\newblock Large scale multi-view stereopsis evaluation.
\newblock In {\em 2014 IEEE Conference on Computer Vision and Pattern
  Recognition}, pages 406--413. IEEE, 2014.

\bibitem{Justin:2020}
Justin Johnson, Nikhila Ravi, Jeremy Reizenstein, David Novotny, Shubham
  Tulsiani, Christoph Lassner, and Steve Branson.
\newblock Accelerating 3d deep learning with pytorch3d.
\newblock In {\em SIGGRAPH Asia 2020 Courses}, SA '20, New York, NY, USA, 2020.
  Association for Computing Machinery.

\bibitem{Kellnhofer:2021:NLR}
Petr Kellnhofer, Lars Jebe, Andrew Jones, Ryan Spicer, Kari Pulli, and Gordon
  Wetzstein.
\newblock Neural lumigraph rendering.
\newblock In {\em CVPR}, 2021.

\bibitem{Kingma:2015:Adam}
Diederik~P. Kingma and Jimmy Ba.
\newblock Adam: {A} method for stochastic optimization.
\newblock In Yoshua Bengio and Yann LeCun, editors, {\em 3rd International
  Conference on Learning Representations, {ICLR} 2015, San Diego, CA, USA, May
  7-9, 2015, Conference Track Proceedings}, 2015.

\bibitem{Klein:2021:ImageIo}
Almar Klein et~al.
\newblock Imageio.
\newblock \url{https://github.com/imageio/imageio}.

\bibitem{Krekel:2004:pytest}
Holger Krekel, Bruno Oliveira, Ronny Pfannschmidt, Floris Bruynooghe, Brianna
  Laugher, and Florian Bruhin.
\newblock pytest.
\newblock \url{https://github.com/pytest-dev/pytest}, 2004.

\bibitem{Laga:2020:SurveyDeepStereo}
Hamid Laga, Laurent~Valentin Jospin, F. Boussaid, and Mohammed Bennamoun.
\newblock A survey on deep learning techniques for stereo-based depth
  estimation.
\newblock {\em IEEE Transactions on Pattern Analysis and Machine Intelligence},
  page 1–1, 2020.

\bibitem{Laine:2020}
Samuli Laine, Janne Hellsten, Tero Karras, Yeongho Seol, Jaakko Lehtinen, and
  Timo Aila.
\newblock Modular primitives for high-performance differentiable rendering.
\newblock {\em ACM Transactions on Graphics}, 39(6), 2020.

\bibitem{Lassner:2020:Pulsar}
Christoph Lassner and Michael Zollh\"ofer.
\newblock Pulsar: Efficient sphere-based neural rendering.
\newblock {\em arXiv:2004.07484}, 2020.

\bibitem{Laurentini:1994:VisualHull}
A. Laurentini.
\newblock The visual hull concept for silhouette-based image understanding.
\newblock {\em IEEE Transactions on Pattern Analysis and Machine Intelligence},
  16(2):150--162, 1994.

\bibitem{Li:2018:Redner}
Tzu-Mao Li, Miika Aittala, Fr{\'e}do Durand, and Jaakko Lehtinen.
\newblock Differentiable monte carlo ray tracing through edge sampling.
\newblock {\em ACM Trans. Graph. (Proc. SIGGRAPH Asia)}, 37(6):222:1--222:11,
  2018.

\bibitem{Lin:2019:PhotometricMeshOptim}
Chen-Hsuan Lin, Oliver Wang, Bryan~C. Russell, Eli Shechtman, Vladimir~G. Kim,
  Matthew Fisher, and Simon Lucey.
\newblock Photometric mesh optimization for video-aligned 3d object
  reconstruction.
\newblock In {\em Proceedings of the IEEE/CVF Conference on Computer Vision and
  Pattern Recognition (CVPR)}, June 2019.

\bibitem{Liu:2020:NeuralSubdivision}
Hsueh-Ti~Derek Liu, Vladimir~G. Kim, Siddhartha Chaudhuri, Noam Aigerman, and
  Alec Jacobson.
\newblock Neural subdivision.
\newblock {\em ACM Trans. Graph.}, 39(4), July 2020.

\bibitem{Liu:2018:Paparazzi}
Hsueh-Ti~Derek Liu, Michael Tao, and Alec Jacobson.
\newblock Paparazzi: Surface editing by way of multi-view image processing.
\newblock {\em ACM Transactions on Graphics}, 2018.

\bibitem{Liu:2019:LearningImplicit}
Shichen Liu, Shunsuke Saito, Weikai Chen, and Hao Li.
\newblock Learning to infer implicit surfaces without 3d supervision.
\newblock In Hanna~M. Wallach, Hugo Larochelle, Alina Beygelzimer, Florence
  d'Alch{\'{e}}{-}Buc, Emily~B. Fox, and Roman Garnett, editors, {\em Advances
  in Neural Information Processing Systems 32: Annual Conference on Neural
  Information Processing Systems 2019, NeurIPS 2019, December 8-14, 2019,
  Vancouver, BC, Canada}, pages 8293--8304, 2019.

\bibitem{Lombardi:2019:NeuralVolumes}
Stephen Lombardi, Tomas Simon, Jason Saragih, Gabriel Schwartz, Andreas
  Lehrmann, and Yaser Sheikh.
\newblock Neural volumes: Learning dynamic renderable volumes from images.
\newblock {\em ACM Trans. Graph.}, 38(4):65:1--65:14, July 2019.

\bibitem{Loubet:2019:RDI}
Guillaume Loubet, Nicolas Holzschuch, and Wenzel Jakob.
\newblock Reparameterizing discontinuous integrands for differentiable
  rendering.
\newblock {\em ACM Trans. Graph.}, 38(6), Nov. 2019.

\bibitem{Luan:2021:Unified}
Fujun Luan, Shuang Zhao, Kavita Bala, and Zhao Dong.
\newblock {Unified Shape and SVBRDF Recovery using Differentiable Monte Carlo
  Rendering}.
\newblock {\em Computer Graphics Forum}, 2021.

\bibitem{Lyu:2020:DiffRefrac}
Jiahui Lyu, Bojian Wu, Dani Lischinski, Daniel Cohen-Or, and Hui Huang.
\newblock Differentiable refraction-tracing for mesh reconstruction of
  transparent objects.
\newblock {\em ACM Transactions on Graphics (Proceedings of SIGGRAPH ASIA
  2020)}, 39(6):195:1--195:13, 2020.

\bibitem{Mescheder:2019:OccupancyNetworks}
Lars Mescheder, Michael Oechsle, Michael Niemeyer, Sebastian Nowozin, and
  Andreas Geiger.
\newblock Occupancy networks: Learning 3d reconstruction in function space.
\newblock In {\em Proceedings IEEE Conf. on Computer Vision and Pattern
  Recognition (CVPR)}, 2019.

\bibitem{Mildenhall:2020:NeRF}
Ben Mildenhall, Pratul~P. Srinivasan, Matthew Tancik, Jonathan~T. Barron, Ravi
  Ramamoorthi, and Ren Ng.
\newblock Nerf: Representing scenes as neural radiance fields for view
  synthesis.
\newblock In {\em ECCV}, 2020.

\bibitem{Nicolet:2021:LargeSteps}
Baptiste Nicolet, Alec Jacobson, and Wenzel Jakob.
\newblock Large steps in inverse rendering of geometry.
\newblock {\em ACM SIGGRAPH Asia 2021}, 2021.

\bibitem{Niemeyer:2020:DVR}
Michael Niemeyer, Lars Mescheder, Michael Oechsle, and Andreas Geiger.
\newblock Differentiable volumetric rendering: Learning implicit 3d
  representations without 3d supervision.
\newblock In {\em Proceedings IEEE Conf. on Computer Vision and Pattern
  Recognition (CVPR)}, 2020.

\bibitem{NimierDavid:2019:Mitsuba2}
Merlin Nimier-David, Delio Vicini, Tizian Zeltner, and Wenzel Jakob.
\newblock Mitsuba 2: A retargetable forward and inverse renderer.
\newblock {\em ACM Trans. Graph.}, 38(6), Nov. 2019.

\bibitem{Oechsle:2021:UNISURF}
Michael Oechsle, Songyou Peng, and Andreas Geiger.
\newblock Unisurf: Unifying neural implicit surfaces and radiance fields for
  multi-view reconstruction.
\newblock In {\em International Conference on Computer Vision (ICCV)}, 2021.

\bibitem{Park:2019:DeepSDF}
Jeong~Joon Park, Peter Florence, Julian Straub, Richard Newcombe, and Steven
  Lovegrove.
\newblock Deepsdf: Learning continuous signed distance functions for shape
  representation.
\newblock In {\em The IEEE Conference on Computer Vision and Pattern
  Recognition (CVPR)}, June 2019.

\bibitem{Paszke:2019:PyTorch}
Adam Paszke, Sam Gross, Francisco Massa, Adam Lerer, James Bradbury, Gregory
  Chanan, Trevor Killeen, Zeming Lin, Natalia Gimelshein, Luca Antiga, Alban
  Desmaison, Andreas Kopf, Edward Yang, Zachary DeVito, Martin Raison, Alykhan
  Tejani, Sasank Chilamkurthy, Benoit Steiner, Lu Fang, Junjie Bai, and Soumith
  Chintala.
\newblock Pytorch: An imperative style, high-performance deep learning library.
\newblock In H. Wallach, H. Larochelle, A. Beygelzimer, F. d\textquotesingle
  Alch\'{e}-Buc, E. Fox, and R. Garnett, editors, {\em Advances in Neural
  Information Processing Systems 32}, pages 8024--8035. Curran Associates,
  Inc., 2019.

\bibitem{Peng:2020:Convolutional}
Songyou Peng, Michael Niemeyer, Lars Mescheder, Marc Pollefeys, and Andreas
  Geiger.
\newblock Convolutional occupancy networks.
\newblock In {\em Computer Vision--ECCV 2020: 16th European Conference,
  Glasgow, UK, August 23--28, 2020, Proceedings, Part III 16}, pages 523--540.
  Springer, 2020.

\bibitem{Ravi:2020:PyTorch3d}
Nikhila Ravi, Jeremy Reizenstein, David Novotny, Taylor Gordon, Wan-Yen Lo,
  Justin Johnson, and Georgia Gkioxari.
\newblock Accelerating 3d deep learning with pytorch3d.
\newblock {\em arXiv:2007.08501}, 2020.

\bibitem{Rockwood:1997:ObjRecFromImages}
Alyn~P Rockwood and Jim Winget.
\newblock Three-dimensional object reconstruction from two-dimensional images.
\newblock {\em Computer-Aided Design}, 29(4):279--285, 1997.
\newblock Reverse Engineering of Geometric Models.

\bibitem{Schoenberger:2016:sfm}
Johannes~Lutz Sch\"{o}nberger and Jan-Michael Frahm.
\newblock Structure-from-motion revisited.
\newblock In {\em Conference on Computer Vision and Pattern Recognition
  (CVPR)}, 2016.

\bibitem{Schoenberger:2016:mvs}
Johannes~Lutz Sch\"{o}nberger, Enliang Zheng, Marc Pollefeys, and Jan-Michael
  Frahm.
\newblock Pixelwise view selection for unstructured multi-view stereo.
\newblock In {\em European Conference on Computer Vision (ECCV)}, 2016.

\bibitem{Schreer:2019:CPVV}
Oliver {Schreer}, Ingo {Feldmann}, Sylvain {Renault}, Marcus {Zepp}, Markus
  {Worchel}, Peter {Eisert}, and Peter {Kauff}.
\newblock Capture and 3d video processing of volumetric video.
\newblock In {\em 2019 IEEE International Conference on Image Processing
  (ICIP)}, pages 4310--4314, 2019.

\bibitem{Schoenberger:2021:COLMAP}
Johannes~Lutz Schönberger et~al.
\newblock {COLMAP} {R}elease 3.6.
\newblock \url{https://github.com/colmap/colmap/releases/tag/3.6}.

\bibitem{Seitz:2006:Survey}
Steven Seitz, Brian Curless, James Diebel, Daniel Scharstein, and Richard
  Szeliski.
\newblock A comparison and evaluation of multi-view stereo reconstruction
  algorithms.
\newblock volume~1, pages 519--528, 01 2006.

\bibitem{Sellan:2021:remesher}
Silvia Sellán and Baptiste Nicolet.
\newblock Libigl botsch-kobbelt local remesher.
\newblock \url{https://github.com/sgsellan/botsch-kobbelt-remesher-libigl},
  2021.

\bibitem{Sitzmann:2019:Siren}
Vincent Sitzmann, Julien~N.P. Martel, Alexander~W. Bergman, David~B. Lindell,
  and Gordon Wetzstein.
\newblock Implicit neural representations with periodic activation functions.
\newblock In {\em Proc. NeurIPS}, 2020.

\bibitem{Starck:2007:SurfaceCapturePerformance}
J. {Starck} and A. {Hilton}.
\newblock Surface capture for performance-based animation.
\newblock {\em IEEE Computer Graphics and Applications}, 27(3):21--31, 2007.

\bibitem{Tamura:2020:TD3M}
R. Tamura, S. Ito, N. Kaneko, and K. Sumi.
\newblock Towards detailed 3d modeling: Mesh super-resolution via deformation.
\newblock In {\em 2020 Joint 9th International Conference on Informatics,
  Electronics \& Vision (ICIEV) and 2020 4th International Conference on
  Imaging, Vision \& Pattern Recognition (icIVPR)}, pages 1--6, Los Alamitos,
  CA, USA, aug 2020. IEEE Computer Society.

\bibitem{Tancik:2020:FourierFeat}
Matthew Tancik, Pratul~P. Srinivasan, Ben Mildenhall, Sara Fridovich-Keil,
  Nithin Raghavan, Utkarsh Singhal, Ravi Ramamoorthi, Jonathan~T. Barron, and
  Ren Ng.
\newblock Fourier features let networks learn high frequency functions in low
  dimensional domains.
\newblock {\em NeurIPS}, 2020.

\bibitem{Tewari:2021:AdvancesNeuralRendering}
Ayush Tewari, O Fried, J Thies, V Sitzmann, S Lombardi, Z Xu, T Simon, M
  Nie{\ss}ner, E Tretschk, L Liu, et~al.
\newblock Advances in neural rendering.
\newblock In {\em ACM SIGGRAPH 2021 Courses}, pages 1--320. 2021.

\bibitem{Thies:2019:NeuralTextures}
Justus Thies, Michael Zollh\"{o}fer, and Matthias Nie\ss{}ner.
\newblock Deferred neural rendering: Image synthesis using neural textures.
\newblock {\em ACM Trans. Graph.}, 38(4), July 2019.

\bibitem{Tola:2012:efficient}
Engin Tola, Christoph Strecha, and Pascal Fua.
\newblock Efficient large-scale multi-view stereo for ultra high-resolution
  image sets.
\newblock {\em Machine Vision and Applications}, 23(5):903--920, 2012.

\bibitem{Vlasic:2009:DynamicShapeCapture}
Daniel Vlasic, Pieter Peers, Ilya Baran, Paul Debevec, Jovan Popovi\'{c},
  Szymon Rusinkiewicz, and Wojciech Matusik.
\newblock Dynamic shape capture using multi-view photometric stereo.
\newblock In {\em ACM SIGGRAPH Asia 2009 Papers}, SIGGRAPH Asia '09, New York,
  NY, USA, 2009. Association for Computing Machinery.

\bibitem{Vu:2012:HighAccMultiView}
Hoang-Hiep Vu, Patrick Labatut, Jean-Philippe Pons, and Renaud Keriven.
\newblock High accuracy and visibility-consistent dense multiview stereo.
\newblock {\em IEEE Transactions on Pattern Analysis and Machine Intelligence},
  34(5):889--901, 2012.

\bibitem{Wang:2021:Neus}
Peng Wang, Lingjie Liu, Yuan Liu, Christian Theobalt, Taku Komura, and Wenping
  Wang.
\newblock Neus: Learning neural implicit surfaces by volume rendering for
  multi-view reconstruction.
\newblock {\em NeurIPS}, 2021.

\bibitem{Wen:2019}
Chao Wen, Yinda Zhang, Zhuwen Li, and Yanwei Fu.
\newblock Pixel2mesh++: Multi-view 3d mesh generation via deformation.
\newblock In {\em ICCV}, 2019.

\bibitem{Wu:2011:HighQualShape}
Chenglei Wu, Bennett Wilburn, Yasuyuki Matsushita, and Christian Theobalt.
\newblock High-quality shape from multi-view stereo and shading under general
  illumination.
\newblock In {\em CVPR 2011}, pages 969--976, 2011.

\bibitem{Yariv:2021:DTUMVS}
Lior Yariv, Yoni Kasten, Dror Moran, Meirav Galun, Matan Atzmon, Basri Ronen,
  and Yaron Lipman.
\newblock {D}erived {DTU} {MVS} {D}ataset.
\newblock
  \url{https://www.dropbox.com/sh/5tam07ai8ch90pf/AADniBT3dmAexvm_J1oL__uoa}.

\bibitem{Yariv:2021:IDRCode}
Lior Yariv, Yoni Kasten, Dror Moran, Meirav Galun, Matan Atzmon, Basri Ronen,
  and Yaron Lipman.
\newblock {IDR} {C}ode.
\newblock \url{https://github.com/lioryariv/idr}.

\bibitem{Yariv:2020:multiview}
Lior Yariv, Yoni Kasten, Dror Moran, Meirav Galun, Matan Atzmon, Basri Ronen,
  and Yaron Lipman.
\newblock Multiview neural surface reconstruction by disentangling geometry and
  appearance.
\newblock {\em Advances in Neural Information Processing Systems}, 33, 2020.

\bibitem{Yu:2020:PixelNerf}
Alex Yu, Vickie Ye, Matthew Tancik, and Angjoo Kanazawa.
\newblock {pixelNeRF}: Neural radiance fields from one or few images.
\newblock In {\em CVPR}, 2021.

\bibitem{Yu:2004:ShapeAndRefl}
Tianli Yu, Ning Xu, and Narendra Ahuja.
\newblock Shape and view independent reflectance map from multiple views.
\newblock In Tom{\'{a}}s Pajdla and Jiri Matas, editors, {\em Computer Vision -
  {ECCV} 2004, 8th European Conference on Computer Vision, Prague, Czech
  Republic, May 11-14, 2004. Proceedings, Part {IV}}, volume 3024 of {\em
  Lecture Notes in Computer Science}, pages 602--616. Springer, 2004.

\bibitem{Zhang:2020:PathDiffRend}
Cheng Zhang, Bailey Miller, Kai Yan, Ioannis Gkioulekas, and Shuang Zhao.
\newblock Path-space differentiable rendering.
\newblock {\em ACM Trans. Graph.}, 39(4), July 2020.

\bibitem{Zhang:2000:ImageBasedMultiRes}
Li Zhang and Steven~M. Seitz.
\newblock {Image-based multiresolution shape recovery by surface deformation}.
\newblock In Sabry~F. El-Hakim and Armin Gruen, editors, {\em Videometrics and
  Optical Methods for 3D Shape Measurement}, volume 4309, pages 51 -- 61.
  International Society for Optics and Photonics, SPIE, 2000.

\end{thebibliography}
}

\clearpage
{
    \renewcommand{\appendixpagename}{Supplementary Material} %
    \begin{appendices}
    \section{Dataset Details}

\subsection{DTU MVS Dataset}

Our evaluation is based on the scripts included in the official DTU MVS dataset release~\cite{DTU:2021:DTUMVS}.
In our experiments, we use a derived dataset~\cite{Yariv:2021:DTUMVS} that includes object masks and was assembled in the context of prior work~\cite{Yariv:2020:multiview, Niemeyer:2020:DVR}. Our implementation expects the input in a different file structure and we will release the converted dataset with our paper.

We are not aware of any copyright or license attached to the DTU dataset. This dataset is widely used in the scientific community.

\subsection{Human Body Dataset}

The human body dataset used in the mesh refinement experiments consist of recordings of two persons, taken in a volumetric capture studio~\cite{Schreer:2019:CPVV}. The subjects are hired actors and they consented in writing to be recorded and to their data being used for research purposes. The consent covers showing the subjects' faces in publications.

\section{Implementation Details}

\figVisualHull

Our implementation is built on top of the automatic differentiation framework PyTorch~\cite{Paszke:2019:PyTorch} and includes code released in the context of previous publications~\cite{Mildenhall:2020:NeRF, Justin:2020, Ravi:2020:PyTorch3d, Sitzmann:2019:Siren}. For remeshing we use a library with Python bindings~\cite{Sellan:2021:remesher}; for differentiable rasterization we use the high-performance primitives by Laine~\etal{}~\cite{Laine:2020}. Additionally, we rely on a variety of other libraries~\cite{Jacobson:2018:igl, Hunter:2007:Matplotlib, Harris:2020:Numpy, Dawson:2021:Trimesh, Klein:2021:ImageIo, Clark:2021:Pillow, Bradski:2000:OpenCV, Krekel:2004:pytest, Dombi:2020:ModernGL}.

In our implementation, we do not assume normalized camera positions. However, we require a rectangular bounding box to normalize the domain to a cube centered at $(0, 0, 0)$ with side length 2.

For 3D reconstruction, we also use the bounding box to construct the initial mesh (Figure~\ref{fig:visual_hull_sup}): we place a grid of points ($32 \times 32 \times 32$) inside the bounding box volume and project the points into each camera image. If a point lies outside any image or mask region, it is removed. We reconstruct a mesh surface from the remaining points with marching cubes.

\section{Experiment Details}

\paragraph{Optimization.}

In the 3D reconstruction experiments, we set the gradient descent step size to $10^{-3}$ for both the mesh vertices and the neural shader. For mesh refinement, we use a step size of $10^{-4}$ for the vertices and $2 \cdot 10^{-3}$ for the shader, progressing the shader faster than the mesh. 

\paragraph{Reconstruction Baselines.}

For \colmap{}~\cite{Schoenberger:2016:mvs, Schoenberger:2016:sfm}, we use the official release 3.6 with CUDA support~\cite{Schoenberger:2021:COLMAP}. Similar to prior work~\cite{Niemeyer:2020:DVR}, we clean the dense point clouds with masks. We also perform trimming after the screened Poisson surface reconstruction (``trim7").

For \idr{}~\cite{Yariv:2020:multiview}, we use the official implementation~\cite{Yariv:2021:IDRCode} and run the reconstruction experiments with camera training, using the \texttt{dtu\_trained\_cameras.conf} configuration.

\paragraph{Runtime Decomposition.}

\tabRuntimeExtended

In the main work, we compare the runtime of one gradient descent iteration of our method to the one of \idr{}~\cite{Yariv:2020:multiview}. The comparison uses data produced by an extensive profiling mode that we implemented for our method and a simpler mode implemented on top of \idr{}. When profiling, we disable all intermediate outputs (\eg{} visualizations). In the case of \idr{}, we profile calls to \texttt{RayTracing.forward} and inside these calls those to \texttt{ImplicitNetwork.forward}, obtaining measurements for \emph{Geometry rendering} and \emph{SDF evaluation}, respectively.

For our method, we record the runtime of most operations during reconstruction. Table~\ref{table:runtime_extended_sup} shows the full decomposition for one sample from the DTU dataset. Computing the shading term of our objective function and computing the gradients with back propagation are the main bottlenecks of our method.

\section{Additional Experimental Results}

\paragraph{Ablation Study of Initial Mesh.}

\figAblationInitialGeometry

\figAblationWeights

\newcommand{\nsaaimage}[1]{\raisebox{-0.5\height}{\includegraphics[width=0.5\columnwidth, trim=40 20 260 10, clip]{#1}}}
\begin{figure*}[tb]
    \centering
    \setlength{\tabcolsep}{3pt}
    \begin{tabular}{ccc}
        \nsaaimage{images/ablation/PE4/owl_experiment_positional_mixed_insets_medres.png} &
        \nsaaimage{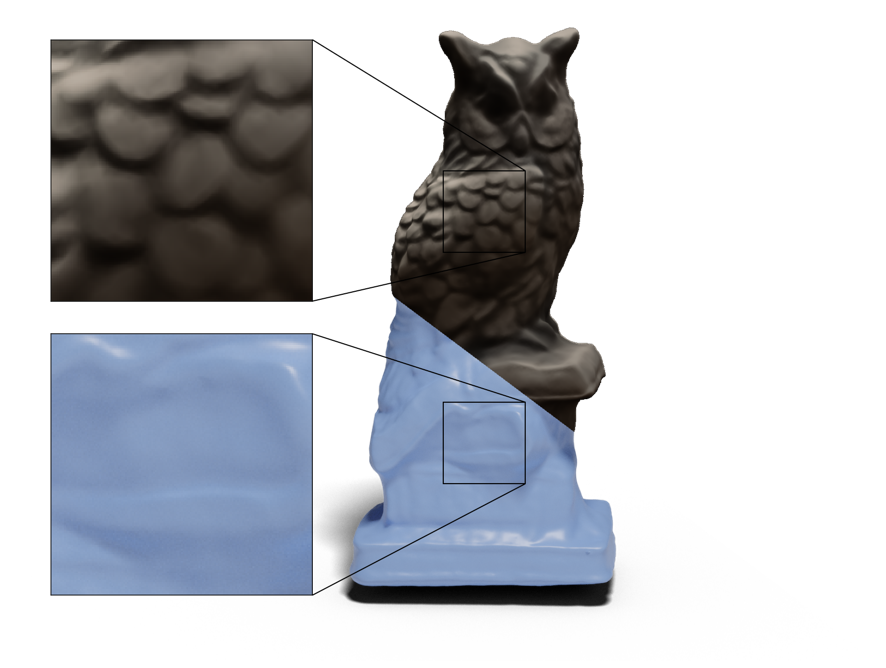} &
        \nsaaimage{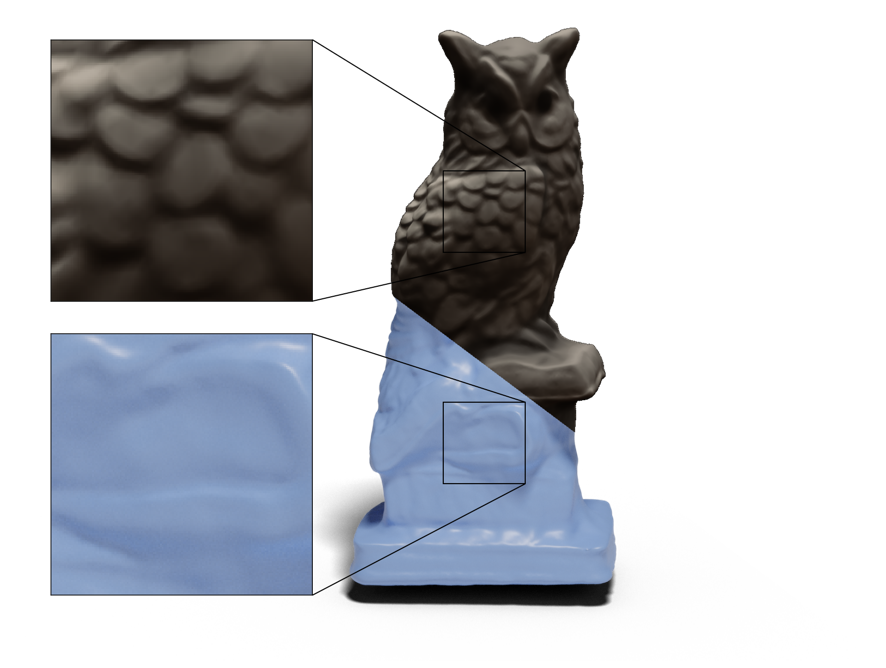} \\
        \small{$3\times256$} & \small{$1\times256$} & \small{$8\times256$} \\
        (Ours) & & \\
        \nsaaimage{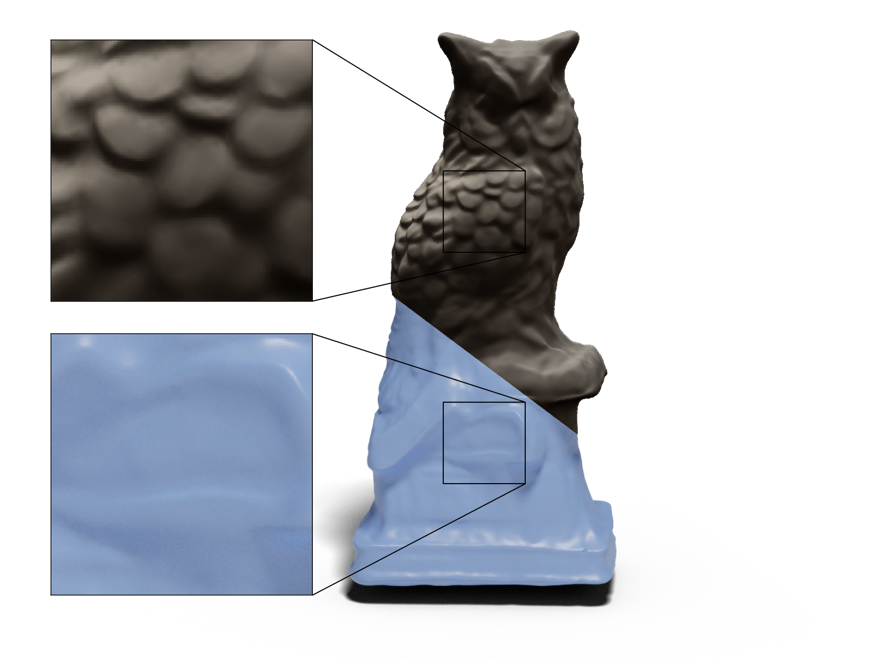} &
        \nsaaimage{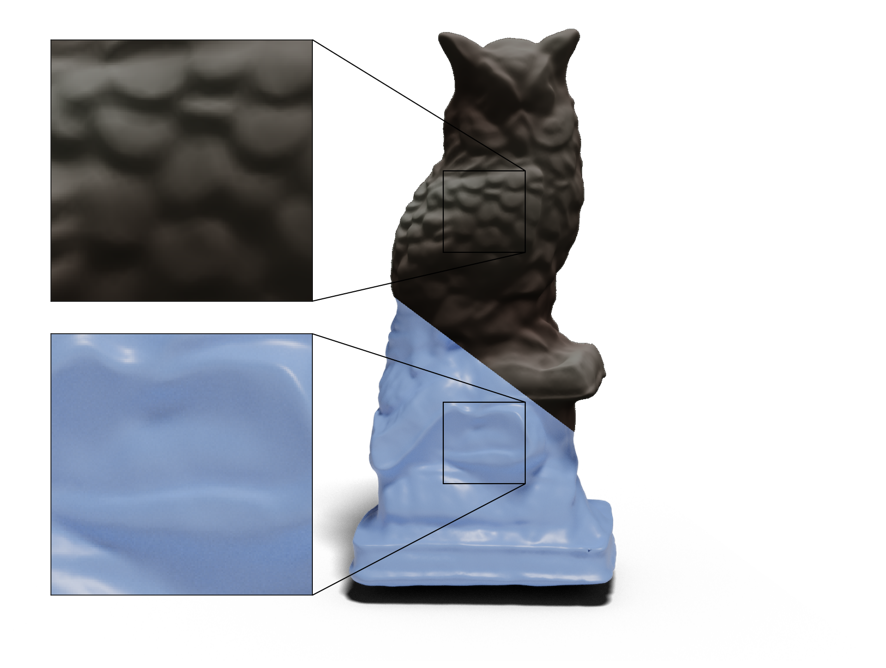} &
        \nsaaimage{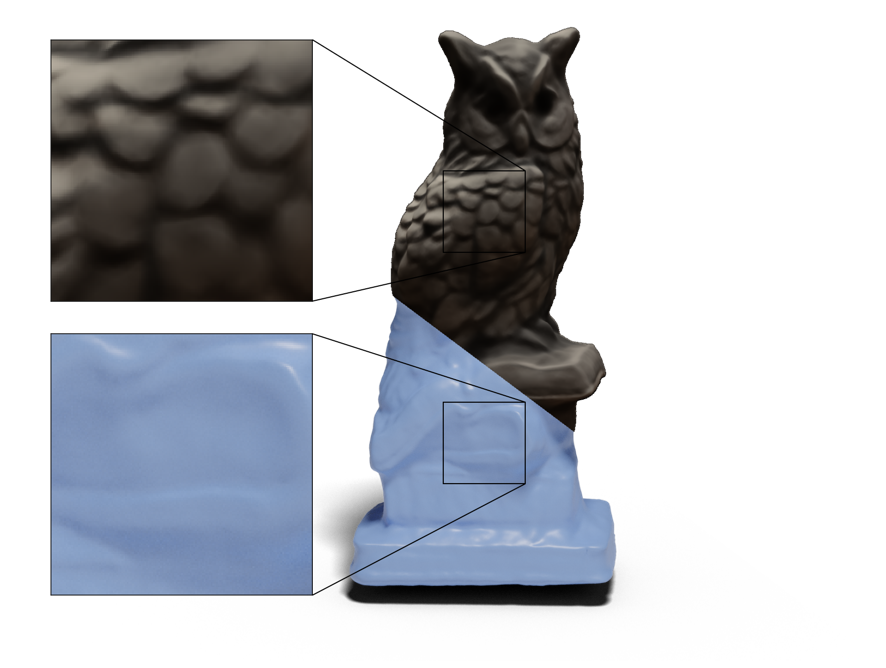} \\
        \small{$16\times256$} & \small{$3\times32$} & \small{$3\times512$} 
    \end{tabular}
    \caption{Different sizes of the positional part of the neural shader (number of layers $\times$ width of layers).}
    \label{fig:different_sizes}
\end{figure*}

We investigate how initial meshes with different resolutions, topology, and geometric distance to the target affect the reconstruction (Figure~\ref{fig:reconstruction_ablation_initial_geometry}). Very coarse initial meshes result in missing details, while fine meshes provide too much geometric freedom, which leads to artifacts. Since we do not support topology changes, holes are only reconstructed if they are present in the initial geometry.

The initial meshes for all DTU objects consist of around 2000 triangles and the output meshes of around 100000 triangles. In terms of scalability, our method handles high-resolution meshes efficiently: in the last 500 iterations (with 2000 iterations in total), the optimization runs on a mesh with the output resolution and still performs fast gradient descent steps.

\paragraph{Ablation Study of Objective Function.}

We investigate the influence of the individual terms of our objective function (Figure~\ref{fig:ablation_objective_function}). Without the Laplacian term, we observe noticeable bumps and crack-like artifacts. 
The normal term has a small influence but improves smoothness around edges. Without the silhouette term, the object boundaries are not properly reconstructed. Without the shading term, the reconstructed shape resembles a smooth visual hull, missing almost all details.

\paragraph{Ablation Study of Network Architecture.}

\figFailureCases

In Figure~\ref{fig:different_sizes}, we show the results for different architecture configurations. Using very few units per layer leads to sharper geometry while the opposite leads to a smoother surface. In the first case, a sharper geometry does not equate with a better estimation of the surface (e.g., shadows are baked into the geometry). In the latter case, we obtain a smooth geometry but we lose geometrical sharpness. Moreover, with the increase in the network parameters, the optimization time grows accordingly.

We found that using 3 layers with 256 units per layers is a good compromise between network complexity and expressive power and yields the best results.

For SIREN~\cite{Sitzmann:2019:Siren}, we noticed that the optimization procedure diverges with our default learning rate. We lowered it to $10^{-4}$ to obtain meaningful results. Despite the lower learning rate, SIREN still converges quickly.

\section{Interactive Viewer}

Since the renderer we use for optimization has the layout of a standard real-time graphics pipeline, the shader part (\ie{} the neural shader) can be readily integrated into other graphics pipelines after training, allowing the interactive synthesis of novel views. 

As an example, we implemented a ``neural" viewer that uses OpenGL to rasterize positions and normals. Then, we use OpenGL-CUDA interoperability and PyTorch to shade the buffers with a pre-trained neural shader. We can envision an exciting extension where the shader is directly compiled to GPU byte code and used similar to other shaders written in high-level shading languages.

\section{Failure Cases}

Figure~\ref{fig:failure_cases_sup} shows some failure cases of our reconstruction method. Since the reconstruction starts from a visual hull-like mesh, deep concavities require large movements in the mesh. This movement is driven by relatively weak shading gradients, which cannot recover these concavities before the mesh becomes overly stiff (\eg{} because the gradient descent step size is reduced during remeshing). On the other hand, silhouette gradients are larger than shading gradients, so the mesh easily ``grows" to fill the masks.

We also observed that the remeshing operation sometimes produces tangled meshes for some objects. This failure case is unrecoverable and the reconstruction needs to be restarted. We are investigating the issue and will coordinate with the authors of the remeshing library. 

Some regions show weak structure in the majority of images for a variety of reasons, for example because they are overexposed. Since we randomly sample one camera view per iteration, there is a high probability to select an image with low structure information for these regions. Using such a view can drive the regions' vertices in unexpected directions away from the actual surface. If such movement occurs at the wrong time (\eg{} right before remeshing), the optimization often fails to correct those vertices.

\section{Ethical Considerations}

Our method has the same potential for misuse as other multi-view 3D reconstruction pipelines. For example, it could be used to digitally reconstruct dangerous objects (\eg{} weapons, weapons parts, ammunition) for reproduction in 3D printing.

Because we also train an appearance model, a malicious actor could potentially modify or edit materials of a reconstructed scene, for example to create fakes by modifying the skin color of a person.

With publicly available code, means for mitigation would be difficult to implement from our side. However, we feel that the overall potential for misuse is low and there are few obvious paths to malicious use or potential damage. Also, our method requires a set of calibrated images, thus is much less accessible than 3D reconstruction or deep fakes based on single images.
    \end{appendices}
}

\end{document}